\documentclass[acmsmall]{acmart}
\AtBeginDocument{%
  \providecommand\BibTeX{{%
    \normalfont B\kern-0.5em{\scshape i\kern-0.25em b}\kern-0.8em\TeX}}}

\setcopyright{acmcopyright}
\copyrightyear{2024}
\acmYear{2024}
\acmDOI{XXXXXXX.XXXXXXX}

\acmJournal{JACM}
\acmVolume{37}
\acmNumber{4}
\acmArticle{111}
\acmMonth{8}

\usepackage{graphicx}
\usepackage{makecell}
\usepackage{multirow}
\usepackage{pifont}
\usepackage{diagbox}
\usepackage{multicol}

\begin{document}

\title{Knowledge Distillation in Federated Learning: a Survey on Long Lasting Challenges and New Solutions}

\author{Laiqiao Qin}
\email{isqlq@outlook.com}
\affiliation{%
  \institution{City University of Macau}
  \streetaddress{Avenida Padre Tomás Pereira Taipa}
  \city{Macau 999078}
  \country{China}}
\author{Tianqing Zhu*}
\email{Tianqing.Zhu@uts.edu.au}
\affiliation{%
  \institution{University of Technology Sydney}
  \streetaddress{123 Broadway}
  \city{Sydney}
  \state{Ultimo NSW 2007}
  \country{Australia}
}

\author{Wanlei Zhou}
\affiliation{%
  \institution{City University of Macau}
  \streetaddress{Avenida Padre Tomás Pereira Taipa}
  \city{Macau 999078}
  \country{China}}
\email{wlzhou@cityu.edu.mo}

\author{Philip S. Yu}
\affiliation{%
  \institution{University of Illinois at Chicago}
  \streetaddress{1200 W Harrison St}
  \city{Chicago}
  \state{Illinois 60607}
  \country{United States}
  }
\email{psyu@cs.uic.edu}

\renewcommand{\shortauthors}{L. Qin, et al.}

\begin{abstract}
Federated Learning (FL) is a distributed and privacy-preserving machine learning paradigm that coordinates multiple clients to train a model while keeping the raw data localized. However, this traditional FL poses some challenges, including privacy risks, data heterogeneity, communication bottlenecks, and system heterogeneity issues. To tackle these challenges, knowledge distillation (KD) has been widely applied in FL since 2020. KD is a validated and efficacious model compression and enhancement algorithm. The core concept of KD involves facilitating knowledge transfer between models by exchanging logits at intermediate or output layers. These properties make KD an excellent solution for the long-lasting challenges in FL.
Up to now, there have been few reviews that summarize and analyze the current trend and methods for how KD can be applied in FL efficiently. 
This article aims to provide a comprehensive survey of KD-based FL, focusing on addressing the above challenges. First, we provide an overview of KD-based FL, including its motivation, basics, taxonomy, and a comparison with traditional FL and where KD should execute. We also analyze the critical factors in KD-based FL in the appendix, including teachers, knowledge, data, and methods. We discuss how KD can address the challenges in FL, including privacy protection, data heterogeneity, communication efficiency, and personalization. Finally, we discuss the challenges facing KD-based FL algorithms and future research directions. We hope this survey can provide insights and guidance for researchers and practitioners in the FL area.
\end{abstract}



\maketitle

\section{Introduction}

Federated Learning (FL) is a distributed collaborative machine learning paradigm to protect data privacy \cite{mcmahan2017communication}. FL enables participants to build a robust and generalized federated model collaboratively through periodic local updates and communication without sharing raw data. Traditional FL algorithms, such as FedAvg \cite{mcmahan2017learning}, involve multiple rounds of client communication for repeatedly averaging local model parameters under server coordination. 
In each round, clients receive the global model from the server to initialize their local model, update their local model using private local data, and upload the updated local model to the server. The server aggregates the received local models to update the global model and then broadcasts the updated model parameters to the clients. The entire process is repeated until the global model converges.

However, this traditional parameter-based paradigm of FL may have some long-lasting challenges. First, sharing model parameters (or gradients) may compromise data privacy \cite{li2020federated,mothukuri2021survey}. For example, adversaries could use these to reconstruct the raw data \cite{geiping2020inverting}. Second, clients' data is mostly not independently and identically distributed (non-IID) \cite{zhao2018federated}, causing local models to suffer from client drift \cite{karimireddy2019scaffold} and deviate from the global optimization objective, resulting in slow convergence \cite{wang2020optimizing} and degraded model performance \cite{zhao2018federated}. Third, the frequent uploading of local models can lead to a severe communication bottleneck \cite{li2020federated,konevcny2016federated,sattler2019robust}. Lastly, parameter-based algorithms lack support for heterogeneous model architectures \cite{li2019fedmd}, personalization for clients \cite{tan2022towards}, and efficient handling of system heterogeneity \cite{ozkara2021quped,kim2021autofl}.

To tackle the above long-lasting challenges in the traditional parameter-based FL is not easy. There are four key reasons that hinder the development of the FL: 1) Since data cannot be shared directly between clients, the clients' model updates lack global awareness~\cite{karimireddy2020scaffold,varno2022adabest}. Traditional FL hopes to solve this problem by aggregating the model parameters of each client. However, 2) model parameters are only the results of client training, not the statistical characteristics of the data. This will lead to deviations in the aggregation model in non-IID scenarios \cite{li2022federated}. 3) The premise of the current mainstream machine learning algorithms is IID, which is often not true in FL \cite{huang2021personalized}. 4) The information in the model parameters is too rich and redundant, far exceeding the minimum amount required for collaborative training, leading to privacy \cite{mothukuri2021survey} and communication \cite{hamer2020fedboost} issues.

\begin{figure}[h]
  \centering
  \includegraphics[width=\linewidth]{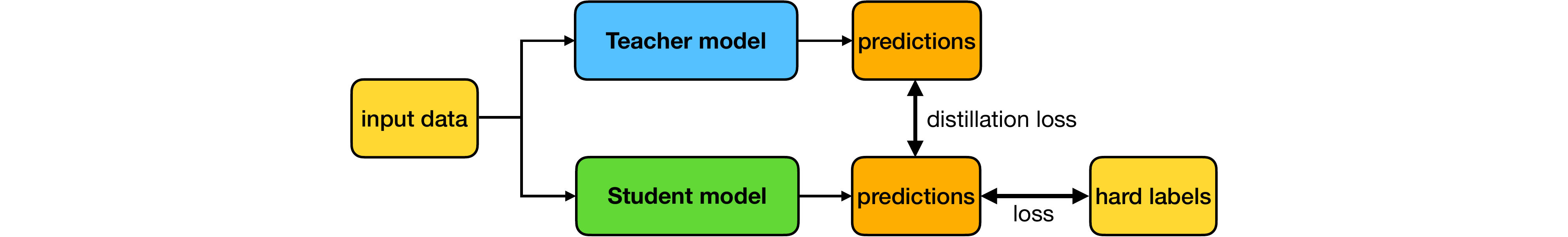}
  \caption{\label{fig:kd}The typical training process of knowledge distillation}
\end{figure}

To fill the gaps, Knowledge distillation (KD) \cite{hinton2015distilling} has been widely applied in FL \cite{lin2020ensemble} since 2020. KD is a highly effective model compression and enhancement technique involving transferring knowledge from a high-performance teacher to a student model \cite{phuong2019towards,mirzadeh2020improved}. Figure \ref{fig:kd} is the basic process of KD, which shows a teacher-student architecture. The teacher network outputs soft labels for the same training data, and the student network uses these soft labels as regularization constraints to learn from the teacher model \cite{hinton2015distilling}. Instead of directly copying the parameters of the teacher model, the student model emulates the soft labels of the teacher model to achieve knowledge transfer. This distinctive approach to transferring knowledge in KD reduces communication costs \cite{sattler2021cfd} and allows for different architectures between teacher and student models\cite{ozkara2021quped}. By the teacher model's soft labels, the student model is endowed with regularization constraints that facilitate superior generalization capabilities.


This typical teacher-student framework in KD is quite suitable for FL. KD-based FL algorithms only require clients to exchange the logits of their local models \cite{chang2019cronus} without uploading model parameters, thus significantly reducing potential privacy risks and communication costs. Unlike parameter-based methods that exchange in the parameter space, KD-based methods exchange in the function space \cite{taya2022decentralized,li2023feddkd}. This allows for the preservation of personalized knowledge between different models to some extent in non-IID scenarios rather than being directly flattened or eliminated \cite{xiao2021novel}. Furthermore, KD is model-agnostic \cite{hinton2015distilling}, and clients have the flexibility to create personalization-focused models with varying architectures\cite{li2019fedmd,huang2023decentralized,jin2022personalized}.

Many algorithmic solutions \cite{bistritz2020distributed,he2020group,horvath2021fjord,ozkara2021quped,zhang2022fine,itahara2021distillation} based on KD have been proposed to tackle the above challenges in FL. These solutions can be roughly divided into three categories: 1) feature-based federated distillation, 2) parameter-based federated distillation, and 3) data-based federated distillation. Feature-based federated distillation algorithms transmit only the model's features between clients and the server, such as logits \cite{jeong2018communication,huang2022learn}, attention \cite{gong2021ensemble,wen2023communication}, and intermediate features \cite{shi2021towards}. Since these features are much smaller and contain far less sensitive information than model parameters, they can effectively address the communication bottleneck \cite{sattler2021cfd} and privacy risks \cite{gong2022preserving} in FL, as well as personalized requirements by using different model architectures \cite{liu2022ensemble}. Parameter-based federated distillation algorithms share model parameters between clients and the server. This method is primarily used to address the issue of model performance degradation caused by directly aggregating model parameters in non-IID scenarios \cite{tang2023fedrad,nguyen2023cadis,peng2023fedgm}. Data-based federated distillation algorithms require clients to use dataset distillation \cite{wang2018dataset} methods to compress their data into a small-scale dataset~\cite{zhou2022communication,song2023federated}, which is not widely adopted in FL and this survey will not delve into this approach further. Table \ref{tab:fd-challenges} summarizes which challenges these three KD-based FL methods are used to solve in FL.

\begin{table}
  \caption{KD-based FL methods are used to solve challenges in FL}
  \label{tab:fd-challenges}
  \begin{tabular}{c|cccc}
    \Xhline{0.08em}
    &Privacy&Non-IID&Communication&Personalization\\ \hline
    \hline
    Feature-based \cite{li2019fedmd} & \checkmark & \checkmark & \checkmark & \checkmark\\
    Parameter-based \cite{shen2022cd2} &  & \checkmark &  &  \\
    Data-based \cite{zhou2020distilled} &  & \checkmark & \checkmark & \\ \hline
  \bottomrule
\end{tabular}
\end{table}

Even though KD-based FL has shown great potential to resolve the long-lasting challenges in FL, the area remains in its early stages, and the associated research exhibits considerable breadth and diversity. The problems and corresponding solutions targeted by these KD-based algorithms are not identical, and the description of "KD in FL" cannot fully summarize them. Different literature has diverse perspectives, explanations, and specific implementations. For instance, while \cite{jeong2018communication} and \cite{li2019fedmd} both use the average value of clients' logits as knowledge, \cite{jeong2018communication} uses local private data, whereas \cite{li2019fedmd} uses public data. Likewise, although \cite{sattler2021fedaux} and \cite{lee2022preservation} are both based on FedAvg, \cite{sattler2021fedaux} uses the local model as a teacher, whereas \cite{lee2022preservation} uses the global model as a teacher. A comprehensive analysis of these methodologies can accelerate further application of this area.

Our primary focus is on the inherent critical factors of KD-based FL algorithms and how KD addresses the long-lasting challenges of FL. The main contribution of this paper is to analyze the fundamental reasons why KD can resolve the long-lasting challenges on FL. We also analyzed some characteristics of using KD in FL and proposed a taxonomy. The details are summarized as follows: 

\begin{enumerate}

\item We provide a systematic analysis of KD-based FL, summarizing the differences between KD-based FL and parameter-based FL algorithm paradigms and analyzing where KD should execute. By comparing the knowledge transfer methods used by different algorithms, we classify KD-based FL algorithms into three categories: feature-based federated distillation, parameter-based federated distillation, and data-based federated distillation.

\item We summarize the critical factors related to state-of-the-art KD-based FL algorithms, including the role of teachers, types of knowledge, dataset acquisition for distillation, and specific methods employed. Details can be found in the appendix. These factors must be considered by all KD-based FL algorithms. By analyzing these factors, we provide important references for future research in KD-based FL.

\item We provide a comprehensive introduction to the application of KD in solving various challenges in FL. These challenges include privacy preservation, mitigating slow convergence and model performance degradation in data heterogeneous scenarios, reducing communication bottlenecks, and addressing personalized model requirements and system heterogeneity issues resulting from variations in client network and hardware resources.

\item We discuss the challenges facing KD-based FL and future research directions.

\end{enumerate}


\section{Background }\label{sec:overviewofflkd}

\subsection{Federated Learning}

FL enables multiple clients to collaboratively train a global model with the coordination of a central server. 
Suppose there are \(N\) independent clients; the matrix \(D_i\) represents the local data of client \(i\) , where \(i\in [1, N]\). Each client has a different local model \(w_i\) and objective function \(f_i(w)\). Each client is given a different weight \(q_i\). The goal of FL is to obtain a generalized global model \(w_g\) through multiple rounds of local training and global aggregation. \(w_i^r\) represents the local model of client \(i\) in the \(r\)-th round of communication, and \(w_g^r\) represents the global model in the \(r\)-th round. The definition and training process of FL is as follows:

\textbf{Definition} \cite{yang2019federated}. In FL settings, client \(i\) aims to achieve the learning objective by iteratively updating their local models through aggregating model results from multiple clients without exchanging or transmitting their local data \(D_i\), by sharing model parameters or outputs \cite{kairouz2021advances}. This process often involves the assistance of a central server, but it is not necessary for certain scenarios or system architectures.



\begin{figure}[h]
  \centering
  \includegraphics[width=\linewidth]{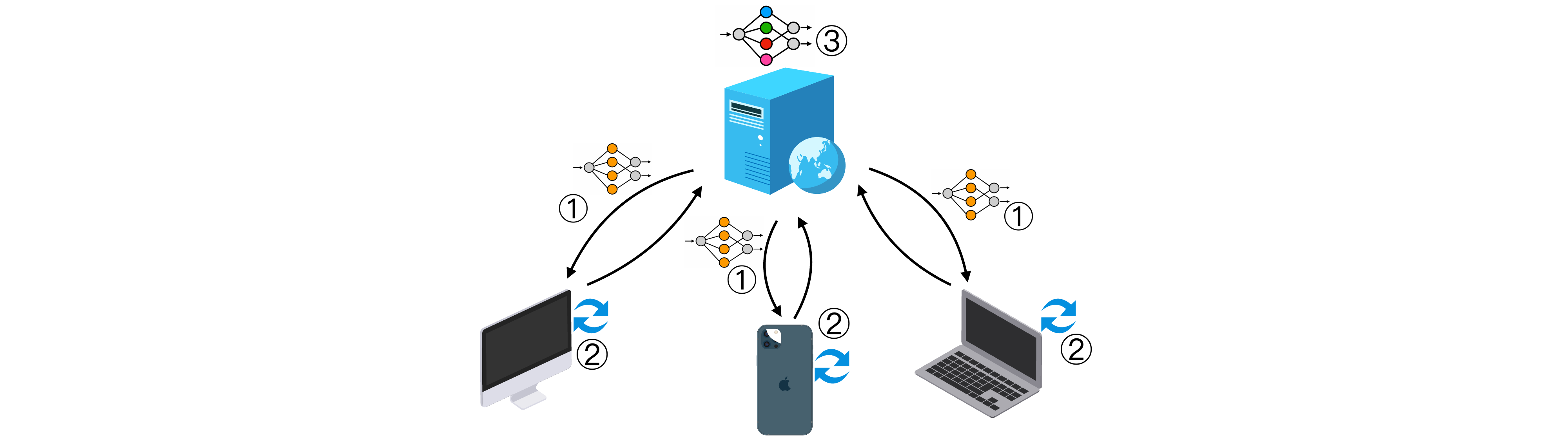}
  \caption{\label{fig:fl}The typical training process of federated learning consists of \ding{172} model distribution, \ding{173} local model update, and \ding{174} global model update.}
\end{figure}

\textbf{Training process}. 
As shown in Figure \ref{fig:fl}, 
the typical training process for FL is as follows:

\begin{enumerate}
\item Model distribution: In round \(r\), the central server distributes the current global model \(w_g^r\) to each client. At the beginning, the global model \(w_g^0\) can be randomly initialized.

\item Local model update: The client \(i\) initializes the local model with the global model \(w_g^r\) and trains it with local data \(D_i\) . When obtaining a new local model \(w_i^{r+1}\), the client sends it to the central server.

\item Global model update: The central server aggregates the local models \(w_i^{r+1}\) to obtain a new global model \(w_g^{r+1}\). Then, the process returns to step 1. The entire process is repeated until convergence.
\end{enumerate}


\subsection{Knowledge Distillation}

\begin{figure}[h]
  \centering
  \includegraphics[width=\linewidth]{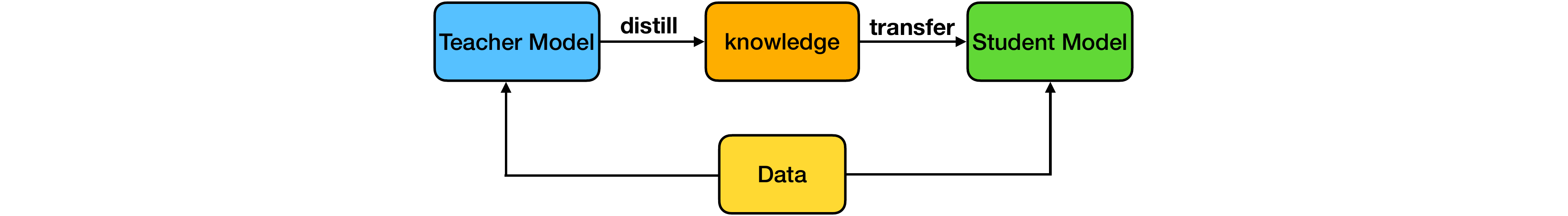}
  \caption{\label{fig:kd-all}KD uses a teacher-student architecture. Teacher model transfers knowledge to the student model.}
\end{figure}

Knowledge distillation (KD) is an effective method for model compression and enhancement \cite{wang2021knowledge}. The main idea is to transfer knowledge from a high-performing teacher model to a student model, as shown in Figure \ref{fig:kd-all}. In KD, knowledge transfer is accomplished by having the student model mimic the soft labels of the teacher model instead of its parameters \cite{phuong2019towards,gou2021knowledge}. This method of knowledge transfer in KD reduces communication costs and permits the student model architecture to differ from that of the teacher model. The soft labels provided by the teacher model serve as regularization constraints for the student model \cite{yuan2020revisiting}, which helps improve its generalization performance.

\textbf{Definition}. In the setting of KD, there are two models: the student model \(S\) and the teacher model \(T\). \(S\) is the model to be trained. By imitating the output of \(T\) on the dataset \(D\), \(S\) can improve its performance. Let \(S_1\) be the model \(S\) trained solely on \(D\) without guidance from \(T\), and \(S_2\) be the model \(S\) trained on \(D\) with guidance from \(T\). Let \(V_t\), \(V_{s_1}\), and \(V_{s_2}\) be the performance measures of \(T\), \(S_1\), and \(S_2\), respectively. Then,

\begin{equation}
\left|V_{s_2}-V_{s_1}\right|\textless\delta,  \left|V_t-V_{s_2}\right|\textless\eta,
\end{equation}
where \(\delta\) and \(\eta\) are non-negative real numbers that satisfy the above equation as small as possible. At this point, we say that \(V_{s_2}\) has a performance gain of \(\delta\), which can be seen as the result of the student model learning from the teacher model. Typically, \(V_t \textgreater V_{s_2}\) due to the capacity difference between the teacher and student model. However, when the complexity of the student model is equivalent to or even exceeds that of the teacher model, \(V_{s_2} \textgreater V_t\) may occur.

\textbf{Training Process}. Assuming there is a pre-trained teacher model T, as shown in Figure \ref{fig:kd}, the typical training process of KD is as follows:

\begin{enumerate}
\item The teacher network outputs soft labels.
\item The student network uses the soft labels from the teacher network as a regularization constraint for the loss function during training. Repeat the entire process until convergence.
\end{enumerate}

Usually use softmax to convert the logits, \(z_i\), into class probabilities, \(q_i\), which are used as soft labels \cite{hinton2015distilling},

\begin{equation}
q_i=\frac{exp(z_i/\tau)}{\sum_j{exp(z_j/\tau)}}
\end{equation}
where temperature \(\tau\) is a hyperparameter, and higher values of \(\tau\) will produce smoother output probability distributions, which will make the model pay more attention to negative labels, and the student model can learn more knowledge from the teacher model \cite{hinton2012neural,phuong2019towards,tang2020understanding}. A special case is to directly match logits, which is equivalent to \(\tau\rightarrow+\infty\).

\subsubsection{Long-lasting Challenges in FL}

Due to its unique setup, FL presents several distinct challenges compared to traditional centralized training.

\textbf{Privacy Preservation}: The original intention of FL was to protect data privacy, but research has shown that sharing only client-side gradients can reveal private data \cite{li2020federated,mothukuri2021survey}. In fact, model parameters such as weights or gradients can be viewed as a "compression" of the dataset, which can be restored using certain techniques \cite{geiping2020inverting,huang2021evaluating}. This creates a conflict where the sharing of intermediate results is necessary to train a global model, but at the same time, sharing these results also poses a risk to data privacy. Thus, there is a need to balance the compression of intermediate results and data privacy.

\textbf{Data Heterogeneity}: An essential assumption in machine learning is that the dataset is independent and identically distributed (IID). However, this assumption does not hold in FL. The datasets of various clients not only differ in size but are also often not identically distributed. Each client may have completely different model updates, causing the local objective to deviate entirely from the global objective, resulting in global knowledge forgetting and client drift \cite{DBLP:journals/corr/abs-2305-19600}. This slows down model convergence and degrades model performance \cite{karimireddy2019scaffold,zhao2018federated,wang2020optimizing}.

\textbf{Communication Bottleneck}: To achieve better model performance, participants in FL often use larger models with improved performance \cite{allen2019convergence,zhou2021over}. However, training a large model collaboratively can increase the communication cost between clients and the server when exchanging model parameters. 
During FL training, there are often many rounds of local training and global aggregation, which require frequent uploading of local models and downloading of global models. In many scenarios, this creates a severe communication bottleneck (e.g., on mobile and edge devices), making FL impractical \cite{konevcny2016federated}.

\textbf{Model Personalization}: For cross-device FL, clients typically use the same model. However, in many cross-silo scenarios, it is a reasonable and practical requirement for each client to design personalized models based on their dataset and specific application scenarios \cite{huang2021personalized,luo2022adapt}. Currently, the mainstream methods in FL are based on parameter-sharing, which does not support heterogeneous models because the knowledge of different models is not compatible in form, and a personalized model cannot directly learn the knowledge of other models through parameters.

\subsection{KD-based FL}

KD-based FL is an emerging paradigm that leverages KD to aggregate the features of client-side models and facilitate collaborative model training. The critical distinction between KD-based FL and parameter-based FL lies in their divergent definitions of knowledge. KD-based FL considers various aspects of the model's features \cite{gou2021knowledge}, such as output layer features \cite{chang2019cronus,li2019fedmd} and intermediate features \cite{gong2021ensemble,yang2022fedmmd}, as knowledge. In contrast, parameter-based FL considers the model's parameters (or model gradients) as knowledge. The two methods exhibit distinct characteristics and face unique challenges within the same privacy protection framework in FL. 

\begin{enumerate}
\item The models communicate through the transmission of dark knowledge \cite{hinton2015distilling}, which is based on features instead of parameters.

\item KD-based FL needs a dataset for distillation, which can be client private data \cite{jeong2018communication}, publicly available data \cite{chang2019cronus}, or artificially generated synthetic data \cite{zhang2022fine}.

\item Typically, KD-based FL lacks a pre-trained teacher model \cite{wang2022knowledge}, and the initial training performance of the teacher model is suboptimal. However, the teacher model gradually improves reliability and convergence as the training progresses.
\end{enumerate}

\begin{figure}[h]
  \centering
  \includegraphics[width=\linewidth]{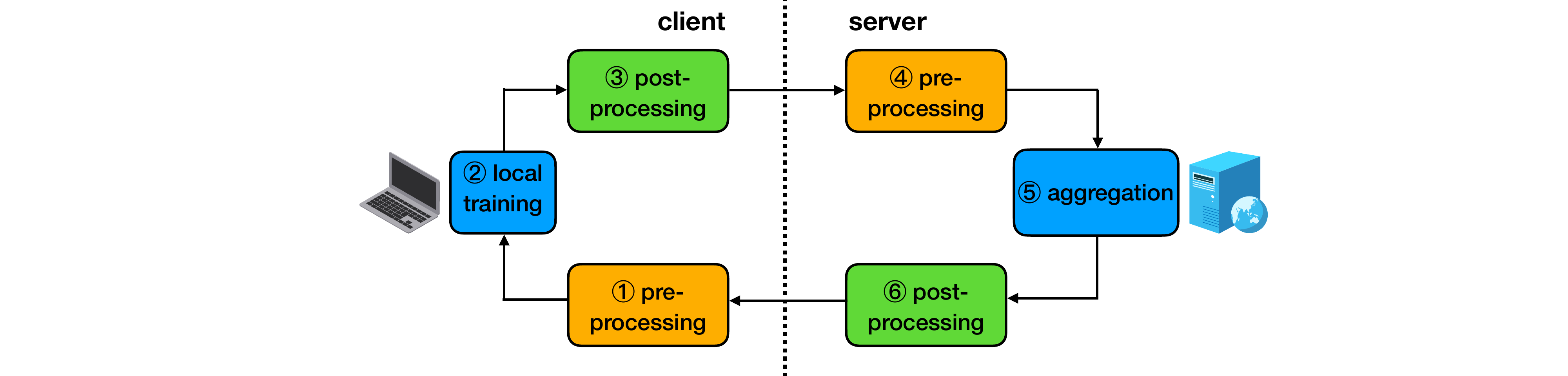}
  \caption{\label{fig:fl-flowchart}The FL process consists of six steps: \ding{172} preprocessing of the global model by clients, \ding{173} local training by clients, \ding{174} further processing of local models by clients, \ding{175} preprocessing of local models by the server upon receiving them, \ding{176} aggregation of local models by the server to obtain the global model, and \ding{177} further processing of the global model by the server. KD can be employed in all six steps.}
\end{figure}

KD-based FL is a concept that lacks a precise definition. Some studies \cite{jeong2018communication,sattler2020communication,ahn2019wireless,xing2022efficient,oh2020mix2fld} refer to specific KD-based FL algorithms as federated distillation (FD), but the meaning of FD varies across papers. In this paper, KD-based FL refers to utilizing KD for addressing specific challenges in FL during local training or server aggregation stages. This paper does not differentiate between KD-based FL and FD concepts to avoid confusion. The section \ref{sec:fdd} will discuss a distinct type of FD that employs dataset distillation \cite{wang2018dataset} instead of knowledge distillation. In this sense, the term FD better represents the discussed concept than KD-based FL.

In the early stages of KD-based FL research, the primary focus was on minimizing communication costs \cite{jeong2018communication,ahn2019wireless}, where clients would upload their local model outputs to the server for aggregation. Subsequently, KD evolved to tackle various challenges in FL and integrated with parameter-based FL. Figure \ref{fig:fl-flowchart} illustrates the comprehensive training process of FL, where KD can be employed at all six steps.

\section{Knowledge-distillation-based Federated Learning}\label{sec:overviewkdfl}


\subsection{KD-based FL: Motivation}

In the context of FL, data cannot leave the local device, and collaborative model training is accomplished through periodic exchange of model information among clients \cite{yang2019federated}. Therefore, the model information exchanged in FL must satisfy specific requirements: 1) the model information must be sufficiently abstract to prevent exposure of raw data privacy, 2) the model information must be sufficiently representative to facilitate the collaborative training of a high-performance generalized model by clients, and 3) the model information must be compact enough to minimize communication costs. Traditional FL algorithms utilize model parameters (or parameter gradients) as the model information \cite{mcmahan2017communication,mcmahan2017learning} to be exchanged among clients to achieve the learning objective of FL.

However, related research \cite{geiping2020inverting,zhu2019deep} has shown that model parameters may not be an ideal model information in collaborative learning. Sharing parameters or gradients may lead to the potential disclosure of sensitive original data \cite{geiping2020inverting,song2020analyzing,huang2021evaluating,lam2021gradient}. Exposing the complete model parameters to potential attackers poses a significant security threat \cite{yin2021comprehensive}. Although some research \cite{mothukuri2021survey,yin2021comprehensive} has been proposed to compensate for the shortcomings of sharing model parameters, inherent defects are challenging to overcome. The concern about privacy in parameter-sharing algorithms is because the information exchanged is too rich and redundant.

In the training process of parameter-based FL, the primary challenge is the communication bottleneck \cite{shahid2021communication,rothchild2020fetchsgd,hamer2020fedboost}. In the context of FL, clients are distributed, and their communication can be unstable, slow, asymmetric, and expensive. Frequent transmission of complete model parameters results in unbearable communication overhead, thereby limiting the participation of potential clients, particularly in scenarios involving mobile or edge devices. Additionally, uploading model parameters slows down communication and prolongs the training process. Moreover, participants may seek to collaborate on training high-performance and large models, significantly augmenting the communication costs. Although some work \cite{luping2019cmfl,sattler2019robust,konevcny2016federated} has been proposed to mitigate communication overhead, overall communication costs are related to model size.

Participants in FL exhibit diverse identities and goals, working collaboratively to achieve their respective objectives. However, parameter-based FL algorithms impose the constraint of utilizing a uniform model across all participants to attain their objectives, which is impractical in real-world scenarios \cite{mansour2020three,kulkarni2020survey}. For instance, in the case of company participation, there may be a desire to acquire a high-performance and large model; however, such a model might not be suitable for certain participants aiming to deploy it on low-performance devices. Parameter-based algorithms fail to address the personalized modeling requirements of clients. Moreover, sharing model parameters fails to address the computational performance and network connectivity disparities arising from system heterogeneity \cite{diao2020heterofl,luo2022tackling}.

The most significant obstacle in FL is data heterogeneity \cite{zhao2018federated,zhu2021federated}. The objective of participant collaboration is to obtain a more robust and generalizable model compared to the one solely trained on local data. The performance of the global model might be compromised by data heterogeneity, resulting in a decline in participants' propensity to collaborate. The stochastic gradient descent method \cite{amari1993backpropagation} is widely employed in machine learning as the primary optimization technique, assuming that the training data follows an independent and identically distributed (IID) pattern. However, the data distribution of each client in FL is often non-IID\cite{zhu2021federated}, resulting in model parameters that solely reflect the optimization outcomes of local private data and fail to capture the optimization results across all data. The optimization directions of different client model parameters exhibit significant variations, and a simple averaging approach fails to yield the globally optimal model. Furthermore, conventional parameter-based algorithms assign different weights to model updates of each client, prioritizing participants with larger datasets while neglecting updates from smaller datasets, leading to fairness concerns in FL \cite{huang2020fairness,li2021ditto}. Additionally, non-IID data distribution presents challenges for model convergence \cite{wu2022node,wang2020optimizing}, particularly in parameter-based algorithms where direct averaging of model parameters fails to capture the underlying knowledge derived from updates contributed by diverse clients.

Therefore, FL urgently requires alternative methods for exchanging model information. KD-based FL adopts KD to exchange the logits of the local model to transmit model information between clients \cite{jeong2018communication,chang2019cronus}. This approach compresses the model information to its minimum, significantly reducing communication overhead and minimizing the sensitive information exposed. Using KD for model enhancement \cite{yuan2020revisiting} enables KD-based FL to address non-IID challenges effectively. Moreover, even for traditional parameter-sharing algorithms, KD is a highly effective supplement that can contribute to client training \cite{he2022class} and server aggregation stages \cite{zhang2022fine}.

\subsection{How KD tackles the challenges of FL}

\subsubsection{Properties of KD}

Due to the unique way of knowledge transfer, KD has some distinct properties that can help to tackle the long-lasting challenges of FL:

\begin{enumerate}
\item Model agnosticism: KD is model-agnostic \cite{tang2020understanding}. The student model can be designed with a personalized architecture based on actual needs without being required to be the same as the teacher model \cite{huang2023decentralized}. This property makes personalization for clients in FL easy to achieve.

\item Model compression: KD was originally proposed to compress models for better production deployment \cite{hinton2015distilling}. In the classic KD architecture, there will be a large model as the teacher network and a small model as the student network. The purpose of training is to obtain a student network as small as possible whose performance can meet actual needs.

\item Capacity gap: The teacher and student models may differ in capacity. The student model may be unable to learn more knowledge from a larger teacher model due to its own limitations \cite{cho2019efficacy}. Therefore, a larger teacher model may not always lead to better distillation results.

\item Model enhancement: The key to model compression is not just reducing model size but also ensuring that a smaller model can still have similar performance \cite{li2020online,cho2019efficacy,phuong2019towards}. Therefore, compressed models need to be enhanced to satisfy actual needs. KD can compress models by transferring dark knowledge while enhancing them.

\item Data dependence: KD is dependent on data \cite{hinton2015distilling}. The teacher model conveys knowledge to the student model by predicting the training data. Therefore, the quality of the training data has a significant impact on the distillation effect. In addition, training data can be unlabeled \cite{menghani2019learning,lee2019overcoming}.

\item Efficient communication: Since only logits need to be transmitted, KD has good communication characteristics in distributed learning scenarios \cite{jeong2018communication}. It enables collaboration between clients without transmitting complete model parameters.
\end{enumerate}

Table \ref{tab:kd-link} puts the properties mentioned above of KD and the challenges in FL to see the relationship between them. These properties give KD an advantage in solving certain challenges in FL.

\begin{table}
  \caption{The relationship between KD properties and FL challenges.}
  \label{tab:kd-link}
  \renewcommand\arraystretch{1.2}
  \resizebox{\linewidth}{!}{
  \begin{tabular}{p{0.16\linewidth}|p{0.215\linewidth}p{0.215\linewidth}p{0.215\linewidth}p{0.215\linewidth}}
    \Xhline{0.08em}
    & Privacy\newline preservation & Non-IID & Communication\newline efficiency & Personalization \\     \hline
    \hline
    Model\newline agnosticism \cite{tang2020understanding}&The client's model architecture is unknown to other clients, helping to protect privacy \cite{gong2022preserving}.&The client can choose an appropriate model architecture based on local data, which helps alleviate non-IID issues \cite{shen2022cd2}.& \multicolumn{1}{c}{\raisebox{-2em}{\centering ---}}  &The model-agnostic nature of KD facilitates customer personalization \cite{afonin2021towards}.\\
    \hline
    Model\newline compression \cite{hinton2015distilling}& \multicolumn{1}{c}{\raisebox{-2em}{\centering ---}} & \multicolumn{1}{c}{\raisebox{-2em}{\centering ---}} &Clients can use compressed small models as intermediary models for transmission \cite{ozkara2021quped}.& \multicolumn{1}{c}{\raisebox{-2em}{\centering ---}} \\
    \hline
    Capacity\newline gap \cite{cho2019efficacy}& \multicolumn{1}{c}{\raisebox{-3em}{\centering ---}} & \multicolumn{1}{c}{\raisebox{-3em}{\centering ---}} & \multicolumn{1}{c}{\raisebox{-3em}{\centering ---}} &When implementing model personalization, the capacity gap needs to be considered to choose an appropriate model architecture.\\
    \hline
    Model\newline enhancement \cite{phuong2019towards}& \multicolumn{1}{c}{\raisebox{-2em}{\centering ---}} &The global model can learn personalized knowledge from the local model, which helps alleviate non-IID issues \cite{zhang2022fine}.& \multicolumn{1}{c}{\raisebox{-2em}{\centering ---}} & \multicolumn{1}{c}{\raisebox{-2em}{\centering ---}} \\
    \hline
    Data\newline dependence \cite{hinton2015distilling}& Knowledge transfer relies on data and may compromise privacy \cite{gong2022preserving}. &Depending on the correlation of the shared distillation dataset to the client dataset, the non-IID problem may be severe or mitigated \cite{liu2022communication}.&There is a communication burden when the shared dataset is large \cite{liu2022communication}.& \multicolumn{1}{c}{\raisebox{-4em}{\centering ---}} \\
    \hline
    Efficient\newline communication \cite{jeong2018communication}& Less information is transmitted, which helps protect data privacy \cite{gong2022preserving}. & \multicolumn{1}{c}{\raisebox{-2em}{\centering ---}} & Only the logits of the model are transmitted to reduce communication costs \cite{jeong2018communication}.& \multicolumn{1}{c}{\raisebox{-2em}{\centering ---}} \\    \hline
  \bottomrule
\end{tabular}
}
\end{table}



Specifically, the approach of KD in perceiving and employing knowledge demonstrates unique potential in addressing challenges within the domain of FL:

\begin{itemize}
\item Privacy preservation: KD does not treat the model parameters as knowledge but the logits of the model as knowledge, effectively reducing privacy concerns associated with the potential leakage of model parameters \cite{geiping2020inverting,huang2021evaluating}. This methodology aligns with the principle of data minimization\cite{biega2020operationalizing}.

\item Data heterogeneity: When dealing with non-IID data, KD distinguishes itself from direct parameter aggregation by utilizing implicit knowledge from diverse client models as mutual regularization constraints. Consequently, this mechanism prevents significant deviations among client models \cite{he2022class}. This regularization constraint is more flexible than parameter aggregation, permitting clients to balance localized and global knowledge \cite{sattler2021cfd}.

\item Communication efficiency: KD maximizes efficiency in knowledge exchange by only sharing the model's outputs among various clients. Consequently, knowledge transmission improved efficacy. Compared to sharing model parameters, distributing model logits significantly reduces communication costs \cite{jeong2018communication,ahn2019wireless}.

\item Personalization: As previously mentioned, KD treats the logits of the model as knowledge, which is independent of the specific model architecture \cite{tang2020understanding}. This characteristic empowers individual clients to design personalized model architectures tailored to their distinct requirements \cite{li2021personalized,ozkara2021quped,yamasaki2023f2mkd}.
\end{itemize}

\subsection{KD-based FL: Taxonomy}

\begin{figure}[h]
  \centering
  \includegraphics[width=\linewidth]{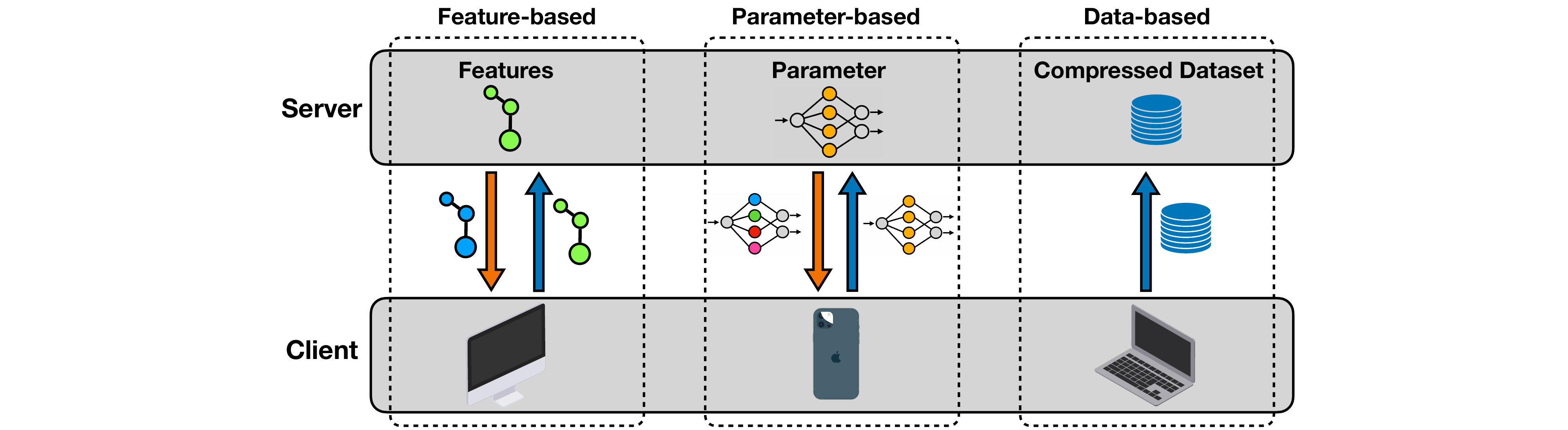}
  \caption{\label{fig:fd-taxonomy}Comparison of the three KD-based FL methods. Feature-based FD shares model features, parameter-based FD shares model parameters, and data-based FD shares local compressed dataset.}
\end{figure}

According to the types of model information shared among clients, federated distillation (FD) can be classified into three categories: feature-based FD, parameter-based FD, and data-based FD. Figure \ref{fig:fd-taxonomy} shows the differences between these three types of FD. Feature-based FD shares model features, parameter-based FD shares model parameters, and data-based FD shares local compressed data. Because the model information exchanged in these three FD methods is different, they exhibit different characteristics when dealing with different challenges in FL. Table \ref{tab:fd-compare} compares the three FDs in dealing with challenges in FL.

\begin{table}
  \caption{Comparison between the three KD-based FL algorithms (feature-based, parameter-based, and data-based) on dealing with challenges in FL}
  \label{tab:fd-compare}
  \renewcommand\arraystretch{1.5}
  \resizebox{\linewidth}{!}{ 
  \begin{tabular}{p{0.22\linewidth}|p{0.26\linewidth}p{0.26\linewidth}p{0.26\linewidth}}
    \Xhline{0.08em}
    & Feature-based \cite{li2019fedmd} & Parameter-based \cite{shen2022cd2} & Data-based \cite{zhou2020distilled} \\    \hline
    \hline
    Privacy & No need to share parameters, less privacy risk \cite{gong2022preserving} & Need to share parameters, which poses privacy risks \cite{geiping2020inverting} & Need to upload local data (compressed), which lead to privacy concerns \\
    \hline
    Non-IID & Clients with heterogeneous data can learn from each other, mitigating non-IID \cite{wen2022communication} & Clients with heterogeneous data can learn from each other, mitigating non-IID \cite{he2022class}  & Can effectively solve non-IID \cite{huang2023federated}\\
    \hline
    Communication & No need to share parameters, low communication cost \cite{jeong2018communication} & Need to share parameters, high communication cost & Communication costs are lower \cite{song2023federated} \\
    \hline
    Personalization & Support heterogeneous models \cite{li2019fedmd} & Heterogeneous models are generally not supported & Heterogeneous models are not supported \\     \hline
    \bottomrule
\end{tabular}
}
\end{table}

\subsubsection{Feature-based FD (No parameter sharing)}

\begin{figure}[h]
  \centering
  \includegraphics[width=\linewidth]{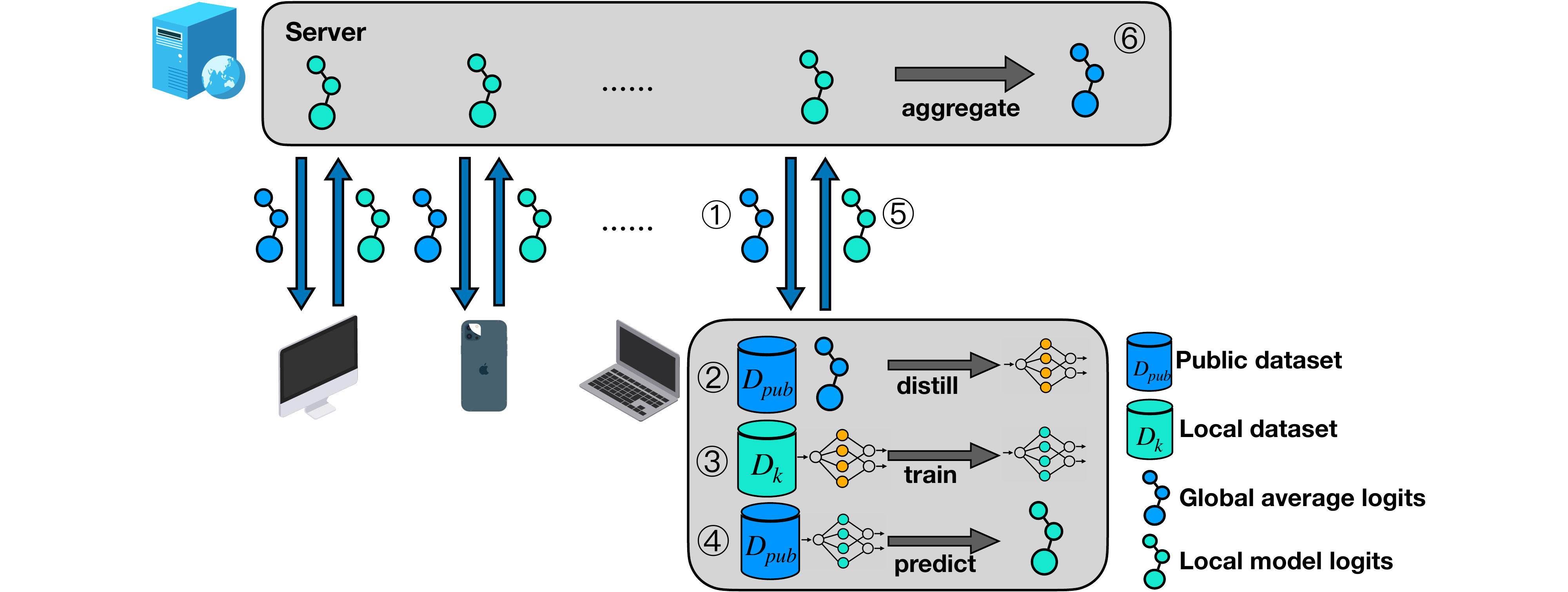}
  \caption{\label{fig:sfd}The typical training process of feature-based FD: \ding{172} clients receive global average logits, \ding{173} clients use global average logits and public dataset to distill into the local models, \ding{174} clients use the distilled models to continue training on the private dataset to obtain the updated models, \ding{175} clients use the updated models to predict on the public dataset and obtain logits, \ding{176} clients upload logits to the server, \ding{177} the server aggregates the logits of all clients to obtain the global average logits.}
\end{figure}

\textbf{Definition}. In feature-based FD, clients share the model features rather than the model parameters, specifically by exchanging either the logits of the model output layer \cite{li2019fedmd} or intermediate representations \cite{gong2021ensemble}. Feature-based FD is a classical algorithm in KD-based FL methods and has received significant attention in recent years due to its remarkable communication efficiency. Figure \ref{fig:sfd} illustrates the typical training process of feature-based FD:

\begin{enumerate}
\item Local logits uploading: Clients train their local models using private data and subsequently submit the logits of these models on a public dataset to the central server.
\item Global logits aggregation: The central server aggregates the logits from all clients and sends the average global logits back to each client.
\item Local model updating: Each client computes the distillation loss by utilizing the global average logits and subsequently updates their local model. Repeat the above steps until the stop condition is satisfied.
\end{enumerate}


\textbf{Advantages}: Feature-based FD leverages the dark knowledge of client models to enhance the efficiency of client communication. Given the universality of this dark knowledge across different model architectures \cite{hinton2015distilling}, clients can easily design heterogeneous model architectures customized to their personalized needs and effectively address challenges arising from system heterogeneity \cite{li2019fedmd,fang2022robust,li2021personalized,he2020group}. The effective knowledge transfer method employed in KD enables clients with heterogeneous data distributions to learn from each other, offering a novel solution for mitigating non-IID problems \cite{ma2021adaptive,ma2021adaptive,huang2021federated,itahara2021distillation}. Additionally, the model's output logits are small, and the communication overheads \cite{jeong2018communication,ahn2019wireless,bistritz2020distributed} and privacy risks \cite{chang2019cronus,kadar2022fedlinked,gong2022preserving} are minimal compared with conventional FL algorithms.

\textbf{Disadvantages}: KD relies on training data for knowledge transfer, and FL lacks access to clients' private data; alternative approaches are often necessary to acquire data for distillation, such as leveraging public datasets \cite{chang2019cronus,li2019fedmd,gong2022preserving,fang2022robust} or employing synthetic data generators \cite{li2021personalized,zhu2021data}. However, the availability of public datasets may be limited in certain scenarios, and collaborative training of generators among clients can impose additional computational. Therefore, exploring simpler and more effective methods for obtaining datasets suitable for distillation is crucial.

\subsubsection{Parameter-based FD (Sharing Parameters)}

\begin{figure}[h]
  \centering
  \includegraphics[width=\linewidth]{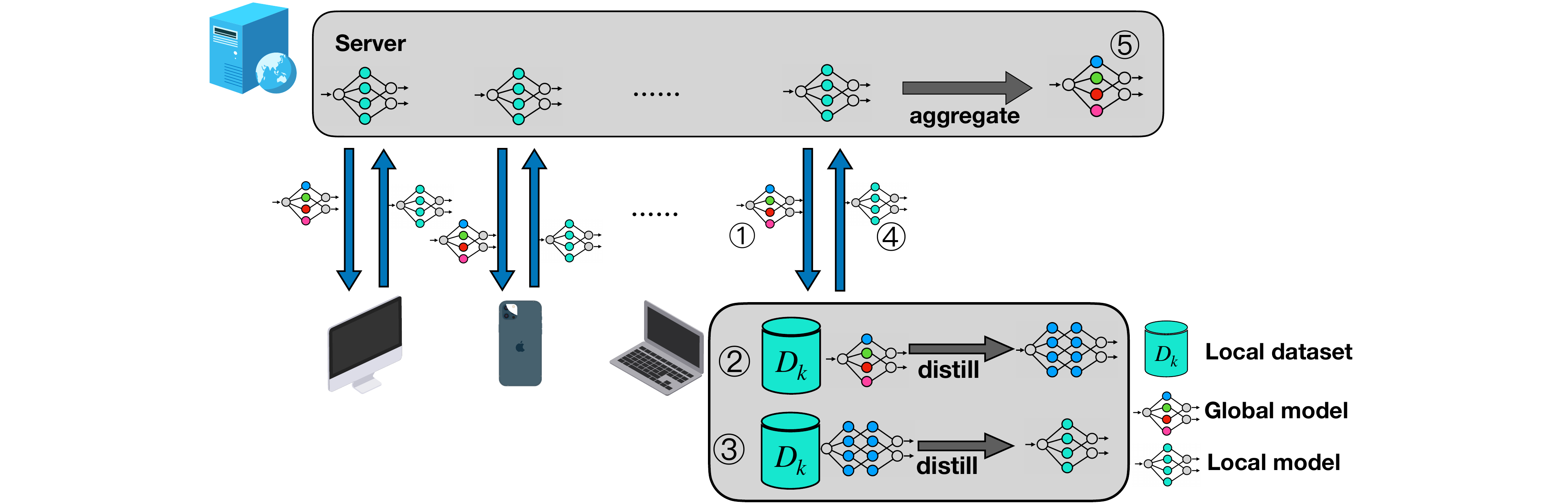}
  \caption{\label{fig:wfd}The typical training process of parameter-based FD: \ding{172} clients receive the global common model, \ding{173} clients use private datasets and the global common model to distill into local private models, \ding{174} clients use private datasets and local private models, and distill into common models, \ding{175} clients upload the updated common models to the server, \ding{176} the server aggregates the common models of all clients to obtain a new global common model.}
\end{figure}

\textbf{Definition}: In parameter-based FD, clients are still required to share model parameters, but KD is utilized to address specific issues during the client training phase \cite{shen2022cd2} or server aggregation phase \cite{zhang2022fine}. This approach extends the capabilities of parameter-based algorithms by enabling more efficient model fusion and updates through the integration of KD. Figure \ref{fig:wfd} illustrates a typical training process of parameter-based FD:

\begin{enumerate}
\item Local common model upload: Clients update the local common models using local private models and private datasets and then upload the common model to the central server.
\item Global common model aggregation: The central server aggregates the common models of all clients to obtain a new global common model and then sends it to each client.
\item Local common model update: Each client uses the global common model to perform KD for the local private model and then uses the updated local private model to distill into the common model.
\end{enumerate}


\textbf{Advantages}: The parameter-based FD algorithm enhances traditional parameter-based FL algorithms. By introducing KD, the parameter-based FD approach can be seamlessly integrated with existing methodologies, demonstrating effectiveness across all stages of FL \cite{lin2020ensemble,chen2020fedbe,liang2022rscfed,shang2022fedic,he2022class}. Instead of simply averaging them, the server can efficiently aggregate diverse local models through dark knowledge fusion. Meanwhile, clients can leverage KD to acquire global knowledge while retaining personalized knowledge.

\textbf{Disadvantages}: Since the clients need to share model parameters, parameter-based FD has a large communication overhead. The process of distillation increases both the computational cost for clients and servers. For instance, achieving model heterogeneity in parameter-based FD settings requires transforming local personalized model architectures into a common architecture through KD \cite{ni2022federated,ozkara2021quped}. Furthermore, parameter-based FD also demands training data for distillation, similar to feature-based FD.

\subsubsection{Data-based FD (Based on dataset distillation)}\label{sec:fdd}

\begin{figure}[h]
  \centering
  \includegraphics[width=\linewidth]{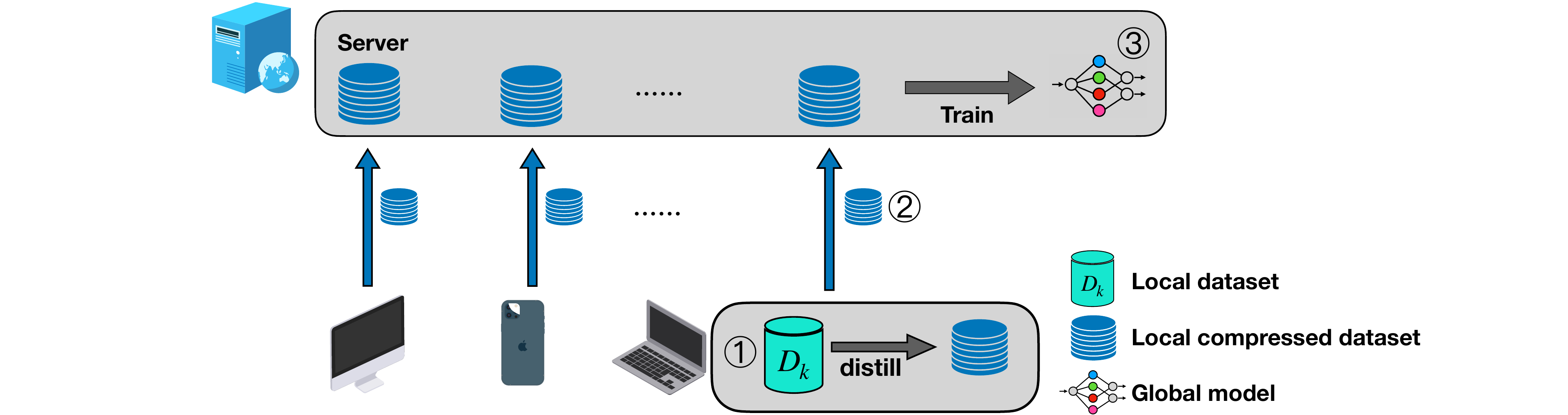}
  \caption{\label{fig:fdd}The typical training process of data-based FD: \ding{172} clients compress local datasets, \ding{173} clients upload compressed datasets, and \ding{174} the server performs centralized training.}
\end{figure}

\textbf{Definition}: In data-based FD, clients employ dataset distillation \cite{wang2018dataset} to compress their local private data, resulting in a small-scale synthetic dataset. This dataset is then transmitted to the server for centralized training \cite{zhou2020distilled,song2023federated}. Unlike traditional FL algorithms, data-based FD does not share model information but rather shares local data (compressed) from the client. To ensure data privacy protection, data-based FD processes the local data so that the original data cannot be recovered. Figure \ref{fig:fdd} illustrates a typical training process of data-based FD:

\begin{enumerate}
\item Local dataset distillation: Clients convert local datasets into small-scale compressed datasets through dataset distillation.
\item Local dataset upload: Clients upload the compressed datasets to the server.
\item Server centralized training: The server performs centralized training using compressed datasets uploaded by all clients.
\end{enumerate}


\textbf{Advantages}: Clients transmit their compressed data directly to the server, facilitating subsequent training on the server without necessitating further client-server communication. This approach ensures efficient communication while addressing non-IID issues by replacing distributed learning with centralized learning \cite{zhou2022communication,xiong2023feddm}.

\textbf{Disadvantages}: The model's performance may decline when trained on the small-scale dataset obtained after dataset distillation \cite{wang2018dataset}, owing to the inherent limitations associated with dataset distillation techniques. Moreover, direct transmission of local data raises significant privacy concerns. Consequently, further investigation is warranted to develop more efficacious and privacy-preserving methodologies for dataset distillation.

\subsection{KD-based FL: Comparison with traditional FL}

This section compares KD-based FL and traditional FL from multiple perspectives, as presented in Table \ref{tab:traditional-fl-compare}.

The traditional FL is characterized by its simplicity in design and applicable to a wide range of client data distributions (suitable for horizontal federated learning, vertical federated learning, and federated transfer learning \cite{yang2019federated}) and scenarios (suitable for cross-device, cross-silo \cite{karimireddy2021breaking}) without requiring additional public datasets. Parameter-based algorithms encompass diverse models in the parameter space, including linear models, tree models, neural network models, and recommendation algorithms \cite{zhang2021survey}. However, the communication overhead associated with parameter sharing is relatively high \cite{luping2019cmfl} and typically scales with the model's size. Since model parameters (weights) represent abstract dataset representations and gradients help identify landmarks for these representations, sharing parameters (or gradients) can lead to privacy leaks and security attacks \cite{geiping2020inverting,lam2021gradient}. Furthermore, parameter sharing is limited to client models with homogeneous architectures.

KD-based FL (specifically, feature-based FD) methods demonstrate a minimal communication overhead, which solely depends on the size of logits and remains unaffected by the model size. As there is no necessity to exchange or transmit model parameters (gradients), the associated privacy risks are notably mitigated. In KD, the student model does not need to be isomorphic to the teacher model, enabling each client's model to possess unique architectures \cite{ozkara2021quped}. However, several relevant algorithms \cite{itahara2021distillation,sun2020federated,gong2021ensemble} based on distillation require clients to download a public dataset, rendering them unsuitable for cross-device scenarios with limited storage capacity. Moreover, performing distillation operations requires additional computational efforts from clients and servers. Typically, distillation-based approaches are better suited for neural network models.

\begin{table}
  \caption{Comparison of KD-based FL and traditional FL}
  \label{tab:traditional-fl-compare}
  \renewcommand\arraystretch{1.2}
  \resizebox{\linewidth}{!}{
  \begin{tabular}{p{0.2\linewidth}|p{0.4\linewidth}p{0.4\linewidth}}
    \Xhline{0.08em}
    & KD-based FL \newline(specifically, feature-based FD) & Traditional FL \\ \hline
    \hline
    Privacy & No need to share parameters, less privacy risk \cite{chang2019cronus} & Need to share parameters, there are privacy risks \cite{yin2021comprehensive} \\
    Non-IID & Clients with heterogeneous data can learn from each other, mitigating non-IID \cite{zhu2021data} & Direct aggregation causes serious non-IID problems \cite{zhu2021federated} \\
    Communication & No need to share parameters, low communication cost \cite{ahn2019wireless}  & Need to share parameters, high communication cost \cite{konevcny2016federated} \\
    Personalization & Support heterogeneous models \cite{li2019fedmd} & Heterogeneous models are not supported \\ \hline
    Public dataset & Usually requires public data sets \cite{li2019fedmd}, or synthetic data \cite{zhu2021data} & No public dataset required \\
    Computing \newline costs & The computational cost is higher due to the additional knowledge distillation process required & Just train the model itself, no additional distillation process required \\ \hline
  \bottomrule
\end{tabular}
}
\end{table}

\subsection{KD-based FL: Where should KD execute}

For an FD system, it may be necessary to consider whether KD should be executed on the client or the server. KD involves two processes: 1) the collection of training data and 2) knowledge transfer. The collection of training data requires a certain storage capacity from the client. KD requires enough training samples to achieve good distillation results \cite{liu2022communication}, which is unsuitable for some client devices with limited storage capacity, such as downloading tens of thousands of images to a mobile phone with small storage space for distillation. The process of knowledge transfer may lead to increased computational overhead, which could pose a burden for devices with lower computing performance, particularly when extracting knowledge from a large model with tens of millions of parameters on low-performance IoT devices. Therefore, an FD algorithm that uses KD to solve the challenges of FL should also consider where KD is executed. 

In FL, KD can be executed on 1) clients or 2) the server. Table \ref{tab:kd-where} compares the advantages and disadvantages of executing KD on the client or server. The execution of KD on clients is generally more suitable for cross-silo \cite{huang2021personalized} scenarios due to the higher computational power of participants' computing devices. However, in the context of cross-device \cite{ur2021trustfed} scenarios, distillation training that needs a substantial amount of additional data and is time-consuming poses challenges due to the limited resources of client devices. When considering KD execution on a server, it is imperative to assess the availability of powerful servers. In traditional FL, the server only performs simple aggregation operations without engaging in resource-intensive tasks. However, if KD is conducted on the server, complexities arise, particularly when mutual distillation among diverse client models needs to be executed \cite{yang2022fedmmd}. This presents a significant challenge for server performance.

\begin{table}
  \caption{Advantages and Disadvantages of executing KD on Client or Server}
  \label{tab:kd-where}
  \renewcommand\arraystretch{1.2}
  \resizebox{\linewidth}{!}{
  \begin{tabular}{p{0.16\linewidth}|p{0.215\linewidth}p{0.215\linewidth}|p{0.215\linewidth}p{0.215\linewidth}}
    \Xhline{0.08em}
    \multirow{2}*{} & \multicolumn{2}{c}{KD on client}  & \multicolumn{2}{c}{KD on the server} \\
    \cline{2-5}
     & Advantages & Disadvantages & Advantages & Disadvantages \\
    \hline
    \hline
    Storage & \multicolumn{1}{c}{\raisebox{-1em}{\centering ---}} & A burden for clients with limited storage capacity & Usually the server does not need to consider storage capacity issues & \multicolumn{1}{c}{\raisebox{-1em}{\centering ---}} \\
    \hline
    Computing\newline performance & Executing KD on the client can reduce the pressure on server computation \cite{he2022class} & High computing performance requirements for clients & Executing KD on the server can reduce the pressure on client computation & Computation bottlenecks occur when performing mutual distillation on large numbers of client nodes \cite{yang2022fedmmd} \\
    \hline
    Communication\newline cost & \multicolumn{1}{c}{\raisebox{-2em}{\centering ---}} & When logits on a public dataset need to be transmitted to the server, the communication cost is substantial \cite{liu2022communication} & No need for the client to transmit logits, and the communication cost is small & \multicolumn{1}{c}{\raisebox{-2em}{\centering ---}} \\ \hline
  \bottomrule
\end{tabular}
}
\end{table}

\section{How KD-based FL tackles the long-lasting challenges}

\begin{figure}[h]
  \centering
  \includegraphics[width=\linewidth]{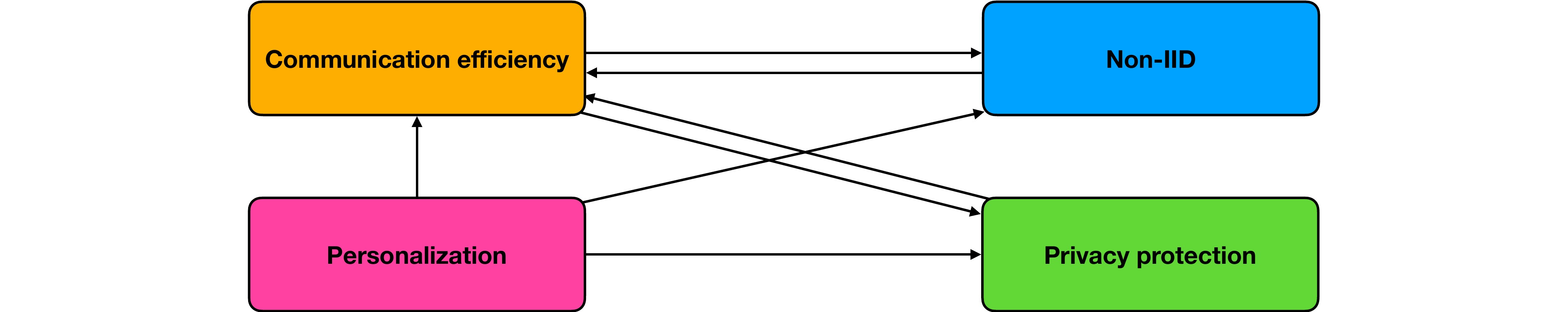}
  \caption{\label{fig:fl-challenge}The relationships between different challenges in FL. Increased communication helps alleviate non-IID but may lead to privacy concerns. Using too many privacy-preserving technologies can lead to a communication burden. Mitigating non-IID helps reduce communication. Personalization helps alleviate communication bottlenecks, privacy concerns, and non-IID.}
\end{figure}

In this section, we will first analyze the relationship between various challenges in FL, including privacy protection, non-IID, communication efficiency, and personalization. In fact, these challenges are not completely independent but interdependent and related. Figure \ref{fig:fl-challenge} shows their relationships. Among them, communication efficiency and privacy protection are positively correlated, and the less information and rounds of communication, the smaller the risk of privacy leakage \cite{chang2019cronus,li2020practical,gong2021ensemble}. There is also a relationship between communication efficiency and non-IID. On the one hand, the non-IID characteristics of client data distribution will slow down the convergence of the model \cite{zhu2021federated,zhu2021data,mills2022client}, resulting in more communication rounds and lower communication efficiency. On the other hand, increasing the number of communication rounds can potentially alleviate the non-IID problem. Model personalization allows different clients to choose personalized model architectures based on their actual needs and local hardware resources, network connections, and other conditions, which can improve communication efficiency and alleviate the negative impact of non-IID \cite{huang2021personalized,jing2023exploring,wu2021personalized}. In addition, model personalization makes the local data structure unknown to other clients, increasing the difficulty of implementing reconstruction attacks. Usually, while solving one problem, other problems are also mitigated. For example, mitigating the non-IID problem will speed up model convergence, reduce communication times, and thereby reduce the risk of privacy leakage.

\subsection{KD-based FL for privacy protection}\label{sec:kd4privacy}


\subsubsection{The overview of privacy protection in KD-based FL}

\paragraph{Why Privacy Challenges Exist}

While conventional FL algorithms seem to guarantee security, recent studies \cite{geiping2020inverting,zhu2019deep,orekondy2018gradient,xu2020information} have revealed that client privacy data can be obtained simply by sharing model parameters or gradients from the client. In fact, sensitive information from the client's training data may be contained in the local weights, aggregated weights, and final model \cite{yin2021comprehensive,xu2020information}. Despite the fact that the raw data is not directly shared, the use of homogeneous models by clients allows the data structure to be known to all participants, making it easy for adversaries to exploit model parameters and gradients to gain additional sensitive information. This can result in severe privacy risks. For instance, in literature \cite{zhao2020idlg}, a method was proposed to extract accurate data from gradients, which is applicable to any differentiable model trained using cross-entropy loss on one-hot labels. In literature \cite{xu2020information}, malicious participants can adjust the training data to cause weight fluctuations in the aggregated model, and privacy information can be obtained by analyzing the trends in weight fluctuations.

\paragraph{Conventional Methods for Privacy Challenges}

Currently, there have been numerous studies \cite{mothukuri2021survey} on privacy leakage in FL. Differential privacy \cite{dwork2006differential} was used in \cite{li2019privacy} to protect client data, but this resulted in reduced model performance. \cite{truex2019hybrid} combined differential privacy with secure multiparty computation \cite{du2001secure}, enabling the federated system to mitigate the impact of noise as the number of participants increased while maintaining the expected level of trust without compromising privacy. However, this increased computation overhead. \cite{yin2021comprehensive} investigated common privacy-preserving mechanisms in FL, including encryption-based privacy-preserving FL \cite{aono2017privacy,chen2020fedhealth}, perturbation-based privacy-preserving FL \cite{hao2019towards,hu2020personalized}, anonymization-based privacy-preserving FL \cite{xie2019slsgd,choudhury2020syntactic}, and hybrid privacy-preserving FL \cite{xu2019verifynet,truex2019hybrid}. These methods are all based on parameter sharing and do not fundamentally address privacy concerns caused by the rich and redundant information carried by model parameters.

\paragraph{Why KD Can Address Privacy Challenge}

In contrast to the rich and redundant information represented by model parameters, KD represents simplicity and efficiency. KD uses a method of sharing the logits of the model to transfer knowledge among clients. Unlike parameter sharing-based methods, in KD, different models provide regularization constraints to other models rather than the models themselves. This regularization constraint is implemented on the same dataset. Typically, in feature-based FD, clients exchange logits generated on the same dataset, greatly reducing communication \cite{chang2019cronus,li2019fedmd}. These logits act as regularizations for the local models, guiding the optimization direction of their training processes without the need to share more details of the local models, thereby protecting the privacy of local data. In \cite{gong2021ensemble}, the predictions and attentions of client models on a public dataset are uploaded to the server to train the global model, reducing communication overhead and protecting privacy. Additionally, for parameter-based FD, although clients still need to share model parameters, privacy protection can still be achieved. For example, \cite{wu2022communication} uses a common model for communication between local personalized models and the global model and transfers knowledge through KD between the local models and the common model, which enhances local data privacy protection to some extent.

\subsubsection{KD-based FL methods for Privacy Challenge}

Currently, some KD-based FL methods have been proposed to enhance data privacy.

In \cite{jeong2018communication}, each client periodically uploads the average logits of local categories (labels) on a private dataset, and the server aggregates the global average logits of each category. Then, each client utilizes the global average logits as the distillation regularization term for local loss. Furthermore, additional local training in \cite{ahn2019wireless} is added based on this approach \cite{jeong2018communication}. By uploading only the average logits of local labels, this method greatly protects the privacy of client data as it does not require uploading model parameters. However, it restricts knowledge sharing among clients, which makes it challenging to ensure the effectiveness of the global model, particularly in non-IID scenarios \cite{jeong2018communication}.

Some research works \cite{chang2019cronus,li2019fedmd,huang2021federated,kadar2022fedlinked} utilize public datasets to achieve more diverse knowledge transfer. Specifically, \cite{huang2021federated} requires each client to upload the logits of its local model on the public dataset and the cross-entropy on its private dataset. After receiving them, the server directly sends them back to each client. Then, each client uses the sum of the mean squared error (MSE) between the logits of other clients and their own logits as the distillation loss to update their local model (where the logits of each client are weighted based on their cross-entropy and similarity to the current client) and then fine-tunes on their local private data. In \cite{kadar2022fedlinked}, each client uploads the logits of their local model on the public dataset to the server, and the server calculates the correlation matrix among clients using these logits. Then, the server uses this matrix to calculate a regularized representation (logits) for each client, which is used for distillation among all clients.

Privacy leaks are usually caused by frequent communication between clients and servers, so many works \cite{gong2022preserving,eren2022fedsplit,zhang2022dense,zhou2020distilled,li2020practical,guha2019one,kang2023one} use the one-shot idea to greatly reduce the number of communication rounds. For example, \cite{li2020practical} divides local data into many parts, each of which is further divided into multiple subsets, and trains a teacher model on each subset. These teacher models vote for the labels of unlabeled public data, obtaining the predicted labels for each public sample. A student model is then trained on the public data with predicted labels. Each client uploads its trained student models for each part to the server, which then uses these student models to predict the labels of each public sample (voting system). Finally, the model is trained on the public data with predicted labels. Although these methods still require uploading student model parameters, they improve communication efficiency and reduce the risk of privacy leakage since only one communication is needed.

In addition, commonly employed privacy protection techniques are utilized in the present study. Specifically, an unlabeled public dataset is used for distillation in \cite{sattler2021fedaux}. During the preparation phase, each client trains a scorer to assign scores to the distillation samples. The training of the scorer incorporates differential privacy \cite{dwork2008differential} to protect client privacy. To enhance privacy protection, differential privacy is applied throughout the entire process in \cite{hoech2022fedauxfdp}, which supplements the approach used in \cite{sattler2021fedaux}. In \cite{gong2022preserving}, quantization and noise integration are implemented on the logits shared among clients to further privacy protection.

\paragraph{Association of Privacy Challenges with Other Challenges}\label{sec:ass-privacy}

Generally speaking, there is a strong correlation between communication efficiency and privacy protection since the greater the amount and frequency of communication exchanges, the higher the likelihood of privacy leakage \cite{gong2022preserving}. Therefore, from a communication perspective, privacy protection requires minimizing the amount of information exchanged and communication frequency as much as possible. One-shot distillation was proposed in \cite{zhou2020distilled,li2020practical,guha2019one,kang2023one}, which limits the number of communication exchanges between clients and servers to once or just a few times, significantly reducing communication costs and privacy risks. Moreover, as illustrated in Figure \ref{fig:fl-challenge}, mitigating non-IID can reduce communication frequency and protect data privacy. Thus, some non-IID mitigation work in FL \cite{gong2022preserving,zhang2022fine,he2022learning,wen2022communication} has also provided some level of privacy protection. Personalization of client models hides the details of local models from other participants, thereby reducing the risk of privacy leakage.

\subsubsection{Discussion of Privacy Protection in KD-based FL}

However, privacy concerns exist in KD-based FL, and the primary concern comes from the data needed for distillation. Since KD depends on data, an extra dataset is typically required for knowledge transfer in FD. Many studies \cite{chang2019cronus,li2019fedmd} use an additional public dataset and assume that it has the same distribution as the clients' local data \cite{li2019privacy}. This assumption can pose a risk of privacy leakage and lead to potential security threats. Some works \cite{zhu2021data,wen2022communication} propose data-free distillation by collaboratively training a generator on the clients to generate synthetic data, eliminating the reliance on a public dataset. However, the process of collaboratively building the generator may still require parameter sharing, leading to privacy risks \cite{zhu2019deep}. Moreover, the synthetic data generated by the generator may also leak clients' privacy data. A potentially feasible method is to ensure that the training data and original privacy data have the same knowledge properties but different representations, such as dataset distillation \cite{wang2018dataset,xiong2023feddm,zhou2020distilled,song2023federated}.

Privacy-preserving techniques commonly used in FL, such as secure multi-party computation \cite{goldreich1998secure}, homomorphic encryption \cite{yi2014homomorphic}, and differential privacy \cite{dwork2006differential}, can still be applied in federated distillation. For example, in \cite{sattler2021fedaux}, differential privacy is used in the local preparation stage to update the local scorers and protect clients' privacy. Building on \cite{sattler2021fedaux}, \cite{hoech2022fedauxfdp} extends the use of differential privacy to the entire federated training process. These privacy-preserving techniques may increase communication overhead and have an impact on model performance. In practical use, finding a balance between privacy, communication, and performance is crucial.

\subsection{KD-based FL for non-IID}\label{sec:kd4non-iid}

\subsubsection{The overview of non-IID in KD-based FL}

\paragraph{Why Non-IID Challenges Exist}

In FL, the data among different clients is usually heterogeneous, which is non-IID \cite{zhu2021federated}. This makes the optimization algorithms used in traditional machine learning and distributed learning unsuitable for FL. Each client uses local data for model updating, and the update directions of each client can be quite different from each other. Direct aggregation of the updates may not represent the globally optimal optimization direction and may even cause the global model to fail to converge. In addition, the amount of data owned by each client may vary significantly, which can result in clients with only a small amount of data having little weight in the aggregation, making the global model unfair. The non-IID in FL can be divided into two dimensions \cite{zhu2021federated}: 1) spatial dimension and 2) temporal dimension. The spatial dimension non-IID is caused by different local data distributions among clients, while the temporal dimension non-IID is caused by different data distributions between new and old samples over time. The former can lead to client drift \cite{mendieta2022local}, while the latter can lead to catastrophic forgetting \cite{shoham2019overcoming}. Although there are conceptual differences between the two, they are both essentially due to the deviation of the optimization direction caused by the difference in data distribution.

\paragraph{Conventional Methods for Non-IID Challenges}

Currently, many studies \cite{zhu2021federated} have been conducted on the non-IID problem in FL. \cite{zhao2018federated} indicates that the performance of the global model will significantly decrease in non-IID scenarios and proposes to use a small subset of data that is globally shared to improve the performance of the global model. \cite{sattler2019robust} proposes a compression framework called "Sparse Ternary Compression (STC)", which extends the top-k gradient sparsity compression technique to achieve a higher downlink communication compression rate in non-IID scenarios. \cite{wang2020optimizing} offsets the model bias caused by non-IID by intelligently selecting appropriate clients to participate in each round of training. \cite{zhu2021federated} summarizes the main approaches to handling non-IID data, including data-based approaches \cite{zhao2018federated,duan2019astraea}, algorithm-based approaches \cite{fallah2020personalized,arivazhagan2019federated}, and system-based approaches \cite{ghosh2020efficient,yin2020fedloc}.

\paragraph{Why KD Can Address Non-IID Challenge}

Based on KD, FD alleviates non-IID problems from different perspectives through a special way of knowledge transmission. The main idea of FD for addressing non-IID problems is to guide models with different optimization directions to mutually influence each other within a certain range through dark knowledge, hoping that local models can learn personalized knowledge from other clients' models and reduce the overfitting problem \cite{nguyen2023cadis}. Similar to how FedProx \cite{li2020federated} constrains the updates of local models during the local training stage, KD-based FL imposes a constraint on local model updates by aggregating the logits of diverse client models, thereby preventing excessive deviation of local models. In addition, KD can keep the local model stable \cite{DBLP:conf/icic/LiuY23a}. Compared with parameter-based FL that directly replaces local models with the global model, which leads to the forgetting of historical personalized knowledge \cite{jin2022personalized}, KD guides local model training in a more gentle way. Moreover, KD is an ideal way to implement model personalization in FL, allowing clients to share knowledge while designing heterogeneous models.

\subsubsection{KD-based FL methods for Non-IID Challenge}

In FL, the most direct solution for client drift \cite{karimireddy2020scaffold} is to learn from other models, including mutual learning \cite{zhang2018deep} between local models or between local and global models. In \cite{wu2023enhancing}, Wasserstein distance \cite{panaretos2019statistical} and regularization terms are introduced into the objective function of federated knowledge distillation to reduce the distribution difference between the global model and client models. \cite{lee2023fedua} proposes a new model aggregation architecture to aggregate models by evaluating the effectiveness of deep neural networks (DNN) and using KD and the uncertainty quantification method of DNN. In \cite{shang2023fedbikd}, knowledge from the global model is used to guide local training to alleviate the local deviation, and local models are used to fine-tune the global model to reduce the volatility of training. In order to deal with heterogeneous tag sets in a multi-domain environment, \cite{wang2023federated} proposes a distillation method based on an instance weighting mechanism to facilitate cross-platform transfer of knowledge. In \cite{xing2022efficient}, a teacher network on the client side teaches a student network using local data, and the student network is sent to the server. The server pairs the student models according to the minimum mean square distance and sends the paired models to each other as their respective teacher networks. In \cite{he2022class}, an auxiliary dataset is used to evaluate the reliability of the global model on certain labels and dynamically adjust the weight of the distillation loss for each class during local training. Specifically, the server aggregates the client models to obtain the global model, and the server calculates the accuracy of the global model for each class using the auxiliary dataset and sends the accuracy to each client. The client optimizes the loss function (hard label cross-entropy + knowledge distillation loss) using the local dataset, where the weight of the KD loss is related to the accuracy of each class. In \cite{yang2022fedmmd}, the clients send their local models to the server, and the server conducts mutual distillation \cite{zhang2018deep} (including features and logits) among all the local models using a labeled public dataset. In \cite{DBLP:conf/icws/ZhengYZYZCD23}, clients use a self-distillation method to train local models. The server generates noise samples for each client and uses them to distill other client models.

Catastrophic forgetting \cite{shoham2019overcoming,ma2022continual,halbe2023hepco} in FL mainly manifests in two aspects: 1) inter-task forgetting and 2) intra-task forgetting. The solutions for both are similar, which is to learn from the old model through KD. In reference \cite{macontinual}, KD is performed on both the client and server sides. On the client side, an unlabeled proxy dataset is used to review the old model, and on the server side, an unlabeled proxy dataset is used to address the non-IID problem among clients. Specifically, a subset of clients is used to learn new tasks, while another subset of clients uses local unlabeled proxy datasets for KD to review old tasks. Client-side distillation is mainly used to address inter-task forgetting, while the server distributes the proxy dataset equally to all clients to obtain soft labels, and then the global model uses these soft labels for KD to address intra-task forgetting. The FL system may experience catastrophic forgetting due to the emergence of new class samples over time. \cite{tang2023fedrad} proposes a method based on relational KD. The local model mines high-quality global knowledge from higher dimensions in the local training stage to better retain global knowledge and avoid forgetting. Reference \cite{dong2022federated} proposes to learn a class-incremental model from a global and local perspective to mitigate catastrophic forgetting. To address local forgetting, the paper designs class-aware gradient compensation loss and class semantic relation distillation loss to balance the forgetting of old classes and extract consistent inter-class relations across tasks. To address global forgetting, the paper proposes a proxy server that assists in local relation extraction by selecting the best old global model.

The performance of the teacher model may be poor during the initial stage, which could potentially misguide the student model. Therefore, \cite{he2022class} uses an auxiliary dataset to evaluate the credibility of the global model. In reference \cite{he2022learning}, this selective learning is further developed in two aspects: 1) the local sample level, i.e., the accuracy of the global model on a certain local sample, and 2) the class level, i.e., the accuracy of the global model on a certain class. Specifically, the server evaluates the global model using the labeled auxiliary dataset to obtain a credibility matrix and sends it to each client. The client uses its local dataset to distill its local model (with the global model as the teacher). During distillation, the local model selectively learns by using a vector obtained from the credibility matrix, which is related to the current sample and relevant classes. The MSE loss function is obtained using this vector matrix as the distillation loss term, and cross-entropy is used as the loss term for the true labels. The two terms are added together to obtain the loss function, and the local model is updated through backpropagation and uploaded to the server.

Non-IID data may cause performance differences in the global model across different clients, leading to fairness issues \cite{li2021ditto,zhou2021towards,wang2021federated}. One approach to address this problem is to design personalized models for local data that participate in global model training through KD while maintaining performance on local data. In literature \cite{ni2022federated}, the global common model is transformed into local personalized models through KD, which is then trained and transformed back to global common models. The personalized models can be uniquely designed based on the client's data distribution and device performance, and the differences between different models can be compensated through deep communication between the models. In literature \cite{kim2022multi}, a novel hierarchical hybrid network is designed, which separates the local and global models into several blocks and connects them at different positions to form multiple paths, similar to a special type of intermediate layer distillation.

Existing FD methods generally do not take into account the diversity of all local models, which can lead to performance degradation of the aggregate model when some local models know little about the sample. In \cite{wang2023dafkd}, the local data of each client is regarded as a specific domain, and a novel domain-knowledge-aware FD method is designed to identify the importance of local models to distillation samples and optimize soft predictions from different models. \cite{DBLP:conf/icassp/ChenCWHZ23} uses entropy to determine the prediction confidence of each local model and select the most confident local model as a teacher to guide the learning of the global model.

\paragraph{Association of Non-IID Challenges with Other Challenges}\label{sec:ass-non-iid}

As mentioned earlier, non-IID can slow down the convergence of the global model, resulting in more communication rounds between clients and the server. However, one feasible way to alleviate non-IID is to increase communication so that heterogeneous clients can have sufficient communication with each other. Of course, this communication is not just about increasing the number of communication rounds but may also involve mutual collaboration between models. In addition, the personalization of models allows clients to design different models based on their local data, which greatly alleviates non-IID problems.

\subsubsection{Discussion of Non-IID in KD-based FL}

Obviously, it is unrealistic to expect a teacher to teach well in an unfamiliar field. The training data used for KD should be relevant to the local data of each client to achieve better results. Many studies \cite{huang2022learn,mo2022feddq,kadar2022fedlinked} assume the existence of a common dataset, which is not realistic in many practical scenarios. Even if a common dataset is available, domain shift \cite{luo2019taking,sankaranarayanan2018learning} may still exist. Therefore, it is a worthwhile direction to study how to transfer knowledge from one domain to another for effective domain adaptation. Some progress has been made in this regard \cite{wang2023federated,wang2023dafkd,su2022domain}. In addition, the compatibility of knowledge between different personalized models is a problem that needs to be carefully considered. Many studies \cite{mi2021fedmdr,ozkara2021quped,bistritz2020distributed,li2019fedmd} use KD to achieve client model personalization based on the assumption that although the architectures of different models are different, the representation of knowledge is the same. This assumption seems reasonable in many FL algorithms, but it lacks more detailed theoretical analysis and proof.

\subsection{KD-based FL for communication efficiency}\label{sec:kd4ce}

\subsubsection{The overview of communication efficiency in KD-based FL}

\paragraph{Why Communication Efficiency Challenges Exist}

To facilitate collaboration among clients, FL requires periodic sharing of model parameters, which can lead to communication bottleneck problems. Communication efficiency is determined by two factors: the amount of information transmitted in each communication round and the number of communication rounds. Unfortunately, the performance of conventional parameter-based FL is inadequate in both aspects. Firstly, sharing complete model parameters (or gradients) in each communication round can result in a large amount of data transmission, which is positively correlated with the size of the model architecture. This can make FL infeasible when using large model architectures. Secondly, the non-IID nature of client data in FL can make it difficult for the global model to converge, which can result in more communication rounds. The decline in communication efficiency can directly lead to a significant prolong of the training time of the global model or even failure. In cross-device scenarios, the frequent joining and exiting of a large number of clients can further exacerbate the situation, leading to additional challenges for FL.

\paragraph{Conventional Methods for Communication Efficiency Challenges}

Currently, numerous studies \cite{shahid2021communication} have investigated communication efficiency issues in FL. Several methods have been proposed to reduce uplink communication costs, such as structured updates and sketched updates \cite{konevcny2016federated}. Additionally, \cite{luping2019cmfl} introduced an orthogonal approach to identify irrelevant client updates, which can help to reduce communication costs. In \cite{sattler2019robust}, a sparse ternary compression framework was proposed to decrease the amount of data transferred in downlink communication. Another approach involves designing a probabilistic device selection scheme \cite{chen2021communication}, which facilitates the selection of high-quality clients in each communication round, as well as using quantization techniques to reduce the amount of communicated data. In \cite{lim2020federated}, common approaches for reducing communication costs were summarized. Although these methods have somewhat reduced the communication costs of downloading or uploading from clients, parameter-sharing-based algorithms generally result in communication costs that are positively correlated with the model size.

\paragraph{Why KD Can Address Communication Efficiency Challenge}

Currently, traditional parameter-sharing FL algorithms require frequent exchange of model parameters, leading to communication bottlenecks. KD uses the model's output instead of the model's parameters as the medium for knowledge propagation, facilitating communication between clients. Compared to model parameters, the model's output is much smaller, greatly reducing communication costs. Classic feature-based FD is mainly used to reduce the amount of data transmitted in a single communication round, not the frequency of communication rounds. Therefore, some one-shot methods \cite{kang2023one,eren2022fedsplit,zhang2022dense,gong2022preserving} have been proposed to reduce the frequency of communication rounds. Additionally, for parameter-based FD, although model parameters still need to be exchanged, communication costs can be greatly reduced because a smaller model can be obtained based on KD for communication between clients \cite{yao2023recognition}.

\subsubsection{KD-based FL methods for Communication Efficiency}

FD was originally proposed to address the problem of the communication bottleneck \cite{jeong2018communication}. Compared with FL algorithms based on parameter sharing, FD greatly reduces communication costs \cite{yang2023communication}. As mentioned earlier, reducing communication costs involves two aspects: 1) reducing the amount of data transmitted in a single communication, and 2) reducing the rounds of communications. We will discuss these two aspects below.

First, FD can greatly reduce the amount of data transmitted in a single communication. \cite{tanghatari2023federated} uses KD to exchange the knowledge learned by the server and the edge device. In order to reduce the communication overhead between the server and the edge device, select the most valuable data to transfer. In \cite{jeong2018communication}, only the average logits of each client's local classes are uploaded, which is negligible compared to the model parameters. Similarly, in \cite{DBLP:conf/iclr/ChenWV23}, clients compute the mean representations of the data classes in their local training dataset and the corresponding mean soft prediction, which is sent to the server for aggregation through differential privacy. However, these methods perform poorly in non-IID settings, such as scenarios where the same label has different features. Therefore, a more common method is to use a public dataset \cite{li2019fedmd} or synthetic data generated by a generator \cite{zhu2021data}. In \cite{sattler2021cfd}, clients use a public dataset for distillation to obtain a distilled model and then use it to train local private data to obtain a local model. Then, the local model predicts soft labels on the public dataset, and the soft labels are quantized, encoded, and sent to the server. The server aggregates the soft labels sent by all clients to obtain the global soft labels and then distills and predicts the global model to obtain soft labels, which are quantized, encoded, and sent to clients. However, regardless of whether the distillation dataset is a public dataset or synthetic data, each client needs to share the output of its local model on the distillation dataset. When dealing with a large distillation dataset, the communication bottlenecks may still be an issue. In \cite{mo2022feddq}, clients use a public dataset for the distillation of local models and then train on local data. Then, they predict soft labels on the public dataset using the local model, as well as compressing the local control variables, which are sent to the server. The server decompresses the compressed data, aggregates soft labels and control variables, and then distills and predicts on the public dataset. Finally, the new soft labels and control variables are compressed and sent to the clients.

In addition, some works focus on reducing the number of communications. In \cite{zhang2022dense}, clients upload their local models to the server, which trains a generator using the integrated models of all clients. The generator then generates synthetic data, and the server uses the average logits of each client's model on the synthetic data (distillation loss) and the global model's cross-entropy loss on the synthetic data to train the model. 

In recent years, a federated learning method based on dataset distillation \cite{wang2018dataset} has been proposed to address communication efficiency issues. Dataset distillation can compress large datasets into smaller ones while maintaining similar model performance. Inspired by dataset distillation, \cite{xiong2023feddm,zhou2020distilled} generate synthetic data for each class based on the parameters distributed by the server and send it to the server. The server uses the synthetic data to update the global model.

\paragraph{Association of Communication Efficiency Challenges with Other Challenges}\label{sec:ass-comm}

FL involves collaboration among multiple clients, which typically relies on network communication and is inherently unstable and uncertain. In traditional centralized training and distributed training in data centers, communication has never been a problem. However, in the context of FL, communication itself can generate many issues that can directly or indirectly impact other aspects. In sections \ref{sec:kd4privacy} and \ref{sec:kd4non-iid}, we respectively discuss the correlation between communication efficiency and privacy protection and non-IID. This correlation can have a ripple effect on the design of FL algorithms. For example, non-IID leads to more communication, which in turn can lead to privacy risks. To address privacy concerns, common cryptographic techniques may be necessary to process communication, which can result in higher communication costs. Moreover, as shown in Figure \ref{fig:fl-challenge}, personalization seems to provide a new solution for heterogeneous client communication. It allows slower clients to choose smaller model architectures, which can improve communication speeds.

\subsubsection{Discussion of Communication Efficiency in KD-based FL}

FD only needs to share logits without sharing model parameters, which seems promising and can greatly reduce communication costs. However, KD relies on data. FD algorithms require clients to share logits for all samples in a shared dataset, and usually, the larger the dataset, the better the performance \cite{liu2022communication}. This can lead to additional communication costs, although these costs are much smaller than directly sharing model parameters. In some scenarios where a common dataset is not available, clients need to collaboratively train a generator to produce synthetic data for distillation, which incurs more computation and communication costs.

\subsection{KD-based FL for personalization}\label{sec:kd4p}

\subsubsection{The overview of personalization in KD-based FL}

\paragraph{Why Personalization Challenges Exist}

In a FL system, clients provide the data, so their own needs should be fully respected. However, in parameter-based FL, clients have difficulty expressing their individual needs because all participants are required to train the same model, even if this model may not be suitable for some participants. This can lead to many problems in practice. On the one hand, participants are likely to refuse to join the FL system because it cannot satisfy their personalized needs, resulting in a smaller number of participants in the FL system and less available training data, which affects the effectiveness of model training. On the other hand, the non-IID problem in FL means that homogeneous models may not have performance advantages for some participants, resulting in fairness issues. In addition, for some clients with lower computational performance, it is difficult to train large federated models, while for some devices with limited bandwidth, it is difficult to bear heavy communication costs. Therefore, personalization is essential for the practical implementation of federated systems. Personalization can be reflected in two aspects: 1) personalization of model architecture and 2) personalization of model size. Clients can design customized model architectures \cite{gad2023federated} based on the specific situations of their local data and the target application fields and can also design models of different sizes based on the computational performance and bandwidth resources of their devices.

\paragraph{Conventional Methods for Personalization Challenges}

Currently, there are many studies \cite{kulkarni2020survey} on personalization in FL. In \cite{arivazhagan2019federated}, a personalized layer method is proposed to alleviate the impact of non-IID. In \cite{deng2020adaptive}, an adaptive personalized FL is proposed, which combines global models and local models into a joint prediction model with an adaptive weight. In \cite{mansour2020three}, three methods for achieving personalization are proposed, including user clustering, data interpolation, and model interpolation. \cite{kulkarni2020survey} and \cite{tan2022towards} summarize commonly used methods for model personalization.

\paragraph{Why KD Can Address Personalization Challenge}

KD is model architecture agnostic, which means that the student model can use an architecture that is completely different from the teacher model. This property provides a natural advantage for achieving model personalization. In feature-based FD, clients can design personalized model architectures according to their own needs, as they only need to share logits that are independent of the model architecture. For parameter-based FD, KD can facilitate knowledge transfer between different model architectures. In this context, KD acts like a converter between models, transforming one into another. Therefore, local models can still be customized, and their knowledge can be transferred to the globally shared model architecture through KD for global aggregation.

\subsubsection{KD-based FL methods for Personalization}

Traditional FL algorithms require all clients to use homogeneous models, while FD allows clients to design heterogeneous models based on their actual needs to achieve personalization. The personalization of the model is mainly reflected in the size and architecture of the model. In conventional FL, clients may be passively or actively excluded due to the limitations of the global model's scalability for local training and deployment or its inability to meet specific business requirements. Different FL settings focus on different personalization, with cross-device scenarios focusing on model size personalization and cross-silo scenarios focusing on model architecture personalization. FD uses the model agnosticism of KD to enable models of different sizes and architectures to exchange knowledge with each other.

\cite{xu2023feddk} propose a serverless framework that uses convolutional neural networks (CNN) to accumulate common knowledge and KD to transfer it. Missing common knowledge is cycled through each federation to provide a personalized model for each group. \cite{su2023federated} introduces a dynamic selection algorithm that utilizes KD and weight correction to reduce the impact of model heterogeneity. \cite{wang2023fedgraph} proposes an effective federated graph learning method based on KD. Each client trains its own local model through KD. In \cite{zhu2022resilient}, a progressive self-distillation \cite{zhang2019your} method is proposed for system heterogeneity and unstable network connections. Simply put, this method divides a large model by column, and each client can download a certain proportion of model parameters based on local network conditions and then supplement the missing parts with local models. Then, the complete local model is updated through progressive self-distillation. Then, based on the network condition, a certain proportion of local model parameters are uploaded. The server aggregates all the incomplete models uploaded by all clients to obtain the global model and broadcasts it to each client. In \cite{chen2022metafed}, a decentralized FL system is designed using cyclic distillation to achieve knowledge accumulation. This method includes two stages: the public knowledge accumulation stage and the personalized stage. Both stages use a circular P2P architecture. In the public knowledge accumulation stage, clients decide whether to use distillation (with the previous client as the teacher) or directly use the previous client's model as the initialization of their local model based on the comparison of the current local model's accuracy and threshold. In the personalized stage, clients use the model and the comparison of the current local model's accuracy and threshold sent by the previous client to determine the weight of the distillation loss term. In \cite{liu2022no}, models of different sizes are assigned to clients with different computing capabilities, with larger models being assigned to stronger clients and smaller models being assigned to weaker clients. A momentum knowledge distillation method is also proposed to better transfer knowledge from the large models of stronger clients to the small models of weaker clients. In \cite{ozkara2021quped,huang2023decentralized}, clients first use KD to learn from the global model locally to train a personalized model. Then, using KD, the personalized model is transformed into a globally common model, which is uploaded to the server for aggregation \cite{ozkara2021quped} or to the other clients for distillation \cite{huang2023decentralized}. \cite{liu2022ensemble} proposes an adaptive quantization scheme for integrated distillation. This method divides clients into different clusters, and local models in the same cluster are isomorphic and have different quantization levels. Model aggregation includes single cluster model aggregation and cluster ensemble distillation loss aggregation.

\paragraph{Association of Personalization Challenges with Other Challenges}

Personalization offers a new approach for solving communication bottlenecks, non-IID, and privacy leakage in FL, as introduced in sections \ref{sec:ass-privacy}, \ref{sec:ass-non-iid}, and \ref{sec:ass-comm}, respectively. In fact, personalization in FL should not be regarded as a challenge but rather as an evolutionary form of FL, namely, an FL that respects differences. A concept related to personalization is system heterogeneity \cite{li2020federated}, which should be considered a challenge in FL. The system heterogeneity challenge refers to the difficulties in designing and implementing an FL system due to the heterogeneity of client devices in terms of computing power and bandwidth resources. Related concepts are system compatibility and scalability \cite{hamood2023clustered}. In some work \cite{chen2023resource,mohammed2023poster,lyu2023prototype}, KD has been used to solve system heterogeneous problems. For example, \cite{chen2023resource} proposes an FL optimization problem based on KD that considers dynamic local resources. This method is used to avoid occupying expensive network bandwidth or bringing a heavy burden on the network. However, the implementation of personalized FL has shifted the focus away from system heterogeneity. Personalization can not only address this problem but also reflect the personalized expectations of clients for the federated system at a higher level rather than just enabling low-performance devices to participate in federated training.

\subsubsection{Discussion of Personalization in KD-based FL}

KD is model-agnostic, but certain drawbacks arise when employing it for personalized FL, which are inherent to the intrinsic characteristics of KD itself. Since KD is data-dependent, achieving personalization requires the sharing of a common dataset globally, which poses a challenge for storage-limited clients. For instance, it is impractical to download a public dataset containing tens of thousands of images into a user's phone for distillation. Moreover, collecting a suitable public dataset is also a challenge, and in practical scenarios, such a dataset may not be available. Additionally, KD increases additional computational costs. Specifically, for parameter-based FD, a KD is required before uploading the local model after receiving the global model, significantly increasing the computational cost for clients and prolonging the training time of the global model.



\section{Future direction}\label{sec:challenges-and-direction}

Recently, KD-based FL has gained popularity as an FL paradigm, and numerous researchers have introduced KD into FL to tackle various challenges. However, KD-based FL faces several challenges. Here are some potential challenges and research directions:
\begin{enumerate}

\item In FL, there is no pre-trained high-performance teacher model \cite{wang2022knowledge,shao2023selective}, and the performance of the teacher model stabilizes gradually during the federated training process. Therefore, the teacher model may mislead the student model. To address this issue, \cite{he2022class} proposes using an auxiliary dataset to evaluate the credibility of the teacher model on specific labels and dynamically changing the weight of the distillation loss on each category during local training. However, this approach requires a labeled public dataset, which limits its practical use. Thus, many studies \cite{fang2022robust,zhuang2023optimizing,itahara2021distillation} use an unlabeled public dataset for distillation, which makes it difficult to evaluate the credibility of the teacher model. In fact, the non-iid characteristics in FL make this problem even more complex, as the teacher model may have better guidance for some clients but may seriously mislead other heterogeneous clients. Therefore, a potential research direction is to design a mechanism to identify teacher model misguidance in a timely manner and control it within a reasonable range.

\item In a typical KD-based FL, each client uploads the logits of the local model output on the public dataset to the server for aggregation to obtain the globally averaged logits, guiding the subsequent training of the local model. This approach is based on the assumption that, although the data from different clients is heterogeneous, the knowledge distilled from this data is homogeneous. Although this assumption has achieved good experimental results in many studies \cite{chang2019cronus,li2019fedmd}, it lacks rigorous theoretical analysis to prove the compatibility of knowledge between different client models and determine the upper bound of this compatibility. Additionally, in non-iid scenarios, the difference between client models is significant, and directly aggregating client logits may lead to global knowledge deviation. Moreover, how to set appropriate hyperparameters in KD to improve learning effects is also an important research direction \cite{alballa2023first}.

\item Many KD-based FL algorithms \cite{zhang2022dense,zhang2022fine,zhu2021data} have proposed data-free distillation to address the challenge of collecting public datasets. The main idea of these algorithms is to use a generator to generate synthetic data for distillation. However, this approach assumes that the data generated by the generator will not expose client privacy, which may not always be the case \cite{xu2019ganobfuscator}. Therefore, addressing the risk of client privacy brought by the generator is a potential research direction. In this regard, specific privacy protection methods and detailed theoretical analysis are needed to determine the extent to which generator-based methods leak privacy and how to balance them with model performance.

\item KD relies on the dataset, and generally, within a certain range, the more data, the better the distillation effect \cite{liu2022communication}. However, adding more data will lead to larger communication costs. Some work \cite{mo2022feddq,gong2022preserving,sattler2021cfd} have explored various methods to reduce communication costs, including quantization and coding. However, these methods still make communication costs proportional to the dataset size. When a large dataset is needed to achieve better distillation results, it still poses a challenge to communication bottlenecks. Therefore, finding a balance between the amount of the public dataset and transmission problems is a worthwhile research direction. Among them, dataset distillation is a potential solution that can compress a massive dataset into a small dataset with equivalent performance. This is very suitable for the FL scenario.

\item Currently, KD-based FL is mostly used in horizontal federated learning scenarios and is rarely used to solve vertical federated learning \cite{yang2019federated}. However, cooperation between enterprises with different feature data is a common requirement in the cross-silo scenario. If knowledge is transferred between clients with different features through exploration in the field of model enhancement using KD, it will solve some of the challenges in model training under vertical federated learning. In this regard, some studies \cite{wang2022vertical,wan2023fedpdd,huang2023vertical} have been conducted.

\item One challenge of using KD for personalized modeling in the FL scenario is that the incentive mechanism is ineffective. The traditional incentive mechanism for parameter-based FL is to calculate client contributions based on their updates. However, in KD-based FL, clients only need to upload logits on the public dataset, which makes it difficult to calculate the true contribution. The server only obtains another set of simplified numbers transformed from the public dataset. Additionally, heterogeneous data on the client side and the potential influence of the teacher model complicate contribution calculations.
\end{enumerate}

\section{Conclusion}\label{sec:conclusion}
KD-based FL is an emerging FL paradigm that provides a series of new solutions to long-lasting challenges in FL by introducing KD. Given the lack of a comprehensive survey on the topic of KD-based FL in the current open literature, this paper provides a detailed survey of the application of KD in FL. First, we introduce the background of FL and KD in detail, especially how some properties in KD can be used to overcome some long-lasting challenges in FL. We then provide an overview of KD-based FL, discussing in detail the motivations, basics, taxonomy, comparison with traditional FL, and where should KD execute. We supplement several critical factors to consider in KD-based FL in the appendix, including the teachers, knowledge, data, and methods. These critical factors are crucial to designing a practical KD-based FL algorithm. We survey and summarize existing KD-based FL methods and divide KD-based FL into feature-based federated distillation, parameter-based federated distillation, and data-based federated distillation. Furthermore, we summarize how KD can be used to solve long-lasting challenges in FL, including privacy, non-IID, communication, and personalization. We also identify some open problems and future directions for further research in this emerging field.

Although current research on KD-based FL is still in its early stages, it is already generating significant interest. KD is expected to become an indispensable component in FL, bringing new possibilities to distributed learning. We hope that this survey paper can provide a useful reference for researchers and practitioners who are interested in KD-based FL and its applications.

\bibliographystyle{ACM-Reference-Format}
\bibliography{main}

\end{document}


\title{Appendix: Knowledge Distillation in Federated Learning: a Survey on Long Lasting Challenges and New Solutions}

\author{Laiqiao Qin}
\email{isqlq@outlook.com}
\affiliation{%
  \institution{City University of Macau}
  \streetaddress{Avenida Padre Tomás Pereira Taipa}
  \city{Macau 999078}
  \country{China}}
\author{Tianqing Zhu*}
\email{Tianqing.Zhu@uts.edu.au}
\affiliation{%
  \institution{University of Technology Sydney}
  \streetaddress{123 Broadway}
  \city{Sydney}
  \state{Ultimo NSW 2007}
  \country{Australia}
}

\author{Wanlei Zhou}
\affiliation{%
  \institution{City University of Macau}
  \streetaddress{Avenida Padre Tomás Pereira Taipa}
  \city{Macau 999078}
  \country{China}}
\email{wlzhou@cityu.edu.mo}

\author{Philip S. Yu}
\affiliation{%
  \institution{University of Illinois at Chicago}
  \streetaddress{1200 W Harrison St}
  \city{Chicago}
  \state{Illinois 60607}
  \country{United States}
  }
\email{psyu@cs.uic.edu}

\renewcommand{\shortauthors}{L. Qin, et al.}

\begin{abstract}
Federated Learning (FL) is a distributed and privacy-preserving machine learning paradigm that coordinates multiple clients to train a model while keeping the raw data localized. However, this traditional FL poses some challenges, including privacy risks, data heterogeneity, communication bottlenecks, and system heterogeneity issues. To tackle these challenges, knowledge distillation (KD) has been widely applied in FL since 2020. KD is a validated and efficacious model compression and enhancement algorithm. The core concept of KD involves facilitating knowledge transfer between models by exchanging logits at intermediate or output layers. These properties make KD an excellent solution for the long-lasting challenges in FL.
Up to now, there have been few reviews that summarize and analyze the current trend and methods for how KD can be applied in FL efficiently. 
This article aims to provide a comprehensive survey of KD-based FL, focusing on addressing the above challenges. First, we provide an overview of KD-based FL, including its motivation, basics, taxonomy, and a comparison with traditional FL. We also analyze the critical factors in KD-based FL, including teachers, knowledge, data, methods, and where KD should execute. We discuss how KD can address the challenges in FL, including privacy protection, data heterogeneity, communication efficiency, and personalization. Finally, we discuss the challenges facing KD-based FL algorithms and future research directions. We hope this survey can provide insights and guidance for researchers and practitioners in the FL area.
\end{abstract}

\keywords{Federated distillation, federated learning, knowledge distillation, privacy preservation, non-IID, communication efficiency, personalization, system heterogeneity}


\maketitle

\section{Critical factors in KD-based FL}
\subsection{KD-based FL: Teachers}

Using diverse models as teachers often means that the problems targeted are different in the FD algorithm. In FD, various teacher models exist, including local and global models, mutual teaching between models, and self-teaching by models themselves. The distinct identities of these teachers often result in varying learning objectives and teaching methods.

\textbf{Local models as teachers}. In FD, a prevalent approach is to consider local models as teacher models and the global model as a student model \cite{gong2022preserving,zhang2022fine,liu2022communication}, which can be seen as a distributed multi-teacher distillation \cite{wu2021one}. Treating local models as teachers offers several advantages. Firstly, the global model can leverage the diverse and comprehensive knowledge from client models to enhance its generalization on unknown samples \cite{zhang2022fine}. Secondly, in the feature-based FD setting, clients' knowledge is presented in a model-agnostic and consistent manner \cite{hinton2015distilling}, facilitating the decoupling of local and global models at the architectural level and freeing local models from constraints imposed by identical architectures \cite{ozkara2021quped}. Moreover, this allows for a better evaluation of each client model's contributions, which is an important consideration in FL systems with incentive mechanisms \cite{li2022incentive}. However, utilizing local models as teachers also entails drawbacks. Firstly, while presenting logits of personalized client models in a unified form may seem feasible, their inner logic might be incompatible due to different representations of the same knowledge during feature extraction across clients \cite{wu2022exploring}. Directly aggregating client model knowledge could potentially lead to performance degradation of the global model. Additionally, clients with higher weight will exert greater influence \cite{zhang2022fine} on the global model's performance, thus making it more dependent on these clients, losing the benefits derived from knowledge diversity offered by multiple clients.

\textbf{Global Model as a Teacher}. In FD, the global model can serve as a teacher model to guide the local models \cite{he2022class,ozkara2021quped,lee2022preservation,yao2021local}, offering several advantages. Firstly, the global model represents a fusion of diverse local models and theoretically exhibits superior generalization capabilities. Consequently, leveraging the global model as a teacher aims to enhance the generalization of local models \cite{lee2022preservation,mo2022feddq}. Secondly, FD facilitates knowledge transfer from the global model to local models through KD in a manner that respects personalization. This personalization is manifested in two aspects: 1) The global model incrementally improves the performance of local models without replacing them, thereby preserving locally personalized knowledge \cite{le2021distilling,ozkara2021quped}, and 2) The architecture of local models can differ from that of the global model, enabling personalization in terms of model architecture \cite{ozkara2021quped}. However, employing the global model as a teacher also entails certain disadvantages. Firstly, clients require additional computational resources for performing KD, which may pose challenges for devices with limited computing capabilities. Secondly, during the initial stages, when the performance of the global model might not be optimal yet, it could misguide local models.

\textbf{Mutual teaching}. Unlike traditional KD, FL typically does not have a pre-trained high-performance teacher model \cite{wang2022knowledge,shao2023selective}. Instead, FL accumulates knowledge from different models through gradual mutual learning \cite{xing2022efficient,zhao2022semi,afonin2021towards,li2022incentive}. Mutual learning between models offers several advantages. Firstly, it mitigates the absence of a high-performance teacher model by enabling models to teach and learn from each other, enhancing generalization performance. Secondly, mutual learning allows models with different distributions to communicate effectively, effectively mitigating the non-IID issue \cite{taya2022decentralized,ma2021adaptive}. Moreover, the exchange of knowledge across diverse models and levels enhances the overall performance of the models. However, one major drawback of mutual learning is its increased computational burden on the entire system and frequent KD that significantly prolongs training time. Some algorithms \cite{yang2022fedmmd} perform mutual learning on servers; however, this approach may lead to the server becoming a bottleneck in the system.

\textbf{Be my own teacher}. Catastrophic forgetting \cite{shoham2019overcoming,shoham2019overcoming,dong2022federated} is a significant challenge in FL, which occurs when a model forgets old knowledge after learning new knowledge, resulting in performance degradation of the original data distribution. Essentially, catastrophic forgetting can be viewed as a non-iid problem in the time dimension. Many FD algorithms \cite{ma2022continual,yao2021local} address this issue by leveraging old models as teachers to guide new models, enabling them to learn new knowledge while retaining previous knowledge. In FL, model forgetting manifests in two main forms \cite{ma2022continual,halbe2023hepco}: 1) inter-task forgetting and 2) intra-task forgetting. Inter-task forgetting \cite{choi2022attractive} means that a new FL task causes the model to forget the knowledge learned in the old task, while intra-task forgetting \cite{ma2022continual} means that the same FL task forgets the knowledge learned in the previous round of communication. To mitigate inter-task forgetting, multiple historical models \cite{yao2021local} are typically employed as teacher models for better review of knowledge across different tasks. For addressing intra-task forgetting, reviewing only the most recent old model \cite{ma2022continual} from previous communication rounds is usually sufficient since early models tend to have lower performance. Generally, local models do not need to review outdated global knowledge because local and global models continue communicating in subsequent updates. However, tackling model forgetting in FL presents certain challenges, such as increased storage costs due to storing historical models; this can be particularly challenging for clients with limited storage resources when multiple historical models are involved. Additionally, reviewing historical models through KD incurs additional computational costs and necessitates retaining old data for improved review effects; however, this may not always be feasible due to privacy concerns and storage cost considerations.

\subsection{KD-based FL: Knowledge}

For the student model, it is imperative not only to identify the teacher model but also to specify which knowledge should be acquired from it. A comprehensive discussion on the topic of the types of knowledge in KD can be found in \cite{gou2021knowledge,chen2021distilling,wang2021knowledge}. In this section, we will examine the characteristics of different types of knowledge under the framework of FL.

For simplicity, we classify knowledge into two categories: simple and complex types. It is evident that compared to simple knowledge, such as logits \cite{hinton2015distilling}, complex knowledge, including intermediate features \cite{romero2014fitnets,qiao2023knowledge,le2023layer} and relationship features \cite{yim2017gift,tang2023fedrad}, offers the student model a more comprehensive understanding. This becomes particularly crucial when addressing the challenge of non-IID in FL systems since complex knowledge can provide diverse insights from different models. However, it should be noted that an abundance of knowledge does not necessarily guarantee improved results; in extreme cases, the teacher model may transfer all its knowledge to the student model, which is equivalent to directly sharing model parameters. In the context of FL, different types of knowledge can influence communication efficiency, computational complexity, model personalization capabilities, and privacy protection.

\textbf{Communication cost}. Communication cost poses a significant challenge in FL, and KD was initially introduced as a solution to the communication bottleneck. Distillation-based algorithms, compared to parameter-sharing-based approaches, effectively mitigate communication costs. In \cite{jeong2018communication}, local class-averaged logits are used as knowledge, which reduces communication costs as the number of classes is typically small. The method, however, demonstrates suboptimal performance in non-iid settings where the same labels have different features. In practice, KD often requires additional public datasets \cite{huang2022learn}, which leads to increased communication costs as the extracted knowledge needs transmission over the network. Although the communication cost is lower than directly transmitting large deep model parameters, it remains non-negligible in scenarios with numerous public data samples and complex knowledge. A certain number of public data samples are necessary for achieving improved distillation results \cite{liu2022communication}, necessitating a trade-off between communication costs and knowledge complexity. Among various types of knowledge \cite{gou2021knowledge}, utilizing output layer logits incurs the lowest communication cost. Nevertheless, tasks involving a large number of categories (e.g., thousands of categories and tens of thousands of public samples) may still encounter communication bottlenecks even when only transmitting logits; thus, compression techniques such as quantization and encoding can be beneficial \cite{sattler2021cfd,gong2022preserving}.

\textbf{Computational cost}. Complex knowledge \cite{romero2014fitnets,yim2017gift,gong2021ensemble,shi2021towards} generally provides more structured and systematic information from the teacher model, making it more beneficial for the student model. However, concerns about both communication and computational resource costs arise in cross-device scenarios. The transfer of complex knowledge can result in increased consumption of computational resources, necessitating the construction of specific data structures \cite{he2022learning,huang2022learn} to facilitate the transfer, thereby further increasing the computational cost of local devices. Hence, it is imperative to balance the performance gains derived from complex knowledge and the associated computational cost.

\textbf{Personalized architecture}. One of the fascinating aspects of FD is that it can easily achieve local model heterogeneity, where different clients can use completely distinct model architectures to satisfy their personalized requirements \cite{wu2019distributed,ozkara2021quped}. Model heterogeneity can lead to diverse internal representations of the same knowledge in various models, so the exchange of complex knowledge, especially the communication between layers and the relationship aspects of different models, becomes more important. Models with different architectures may differ significantly, not only in terms of size but also in complexity. The situation becomes more complex when there are many heterogeneous clients in the FL system. For example, various heterogeneous local models in the server guide the global model \cite{zhang2022dense,gong2022preserving}. At this time, due to the differences in model architecture, it is challenging to achieve compatibility with complex knowledge from various models. One potential solution involves clients converting their personalized model architecture into a common one \cite{sun2022fed2kd} and designing a transmission process for complex knowledge exchange. In practice, heterogeneous models typically utilize simple knowledge for convenience.

\textbf{Privacy protection}. Privacy risk is a crucial concern to be taken into account. Generally, the more information shared, the higher the likelihood of privacy leakage. The logits generated by the model may encompass statistical information from the client \cite{gong2022preserving}, while the intermediate and relationship features \cite{gong2021ensemble} may contain structured and systematic client-specific details. Hence, privacy preservation should also be considered when selecting different types of knowledge. To protect privacy, some techniques, such as differential privacy \cite{hoech2022fedauxfdp,sattler2021fedaux} or quantization \cite{gong2022preserving,sattler2021cfd}, are commonly employed to protect the model's knowledge.

\subsection{KD-based FL: Data}\label{sec:kdfldata}

Like the human learning process, teaching needs textbooks, and KD also relies on training data. In this section, we will comprehensively analyze the training data within the context of FD. FD imposes specific requirements on training data utilized in KD, encompassing three aspects: 1) the distribution of training data should be relevant to the local data of each client \cite{liu2022communication}, 2) there should be sufficient training data  \cite{liu2022communication,park2023towards}, and 3) selecting appropriate data types \cite{jeong2018communication,li2019fedmd,zhu2021data}.

\subsubsection{Correlation}
The effectiveness of KD depends on the correlation between the data used for distillation and the local data on each client \cite{liu2022communication}. For example, using a fruit dataset to guide a student model for cat-dog classification would not yield satisfactory results. KD necessitates that both the teacher and student models employ the same dataset for distillation. However, in the setting of FL, private data cannot leave the client, and the teacher and student models may not be on the same device, making it difficult for the client to use its private data for distillation directly. Therefore, numerous FD algorithms \cite{li2019fedmd,sun2020federated,cho2021personalized,itahara2021distillation} assume access to a correlated public dataset that does not need an identical distribution to the client's data, such as using a cross-domain public dataset for distillation \cite{gong2022preserving}. However, it is difficult to find a relevant public dataset in many FL tasks. Therefore, some data-free FL methods have been proposed in \cite{zhu2021data,zhang2022fine,zhang2022feddtg}, where clients collaborate to train a generator for synthesizing data. However, certain tasks like object detection \cite{zhao2019object} pose challenges when generating synthetic samples using a generator. Moreover, the requirement of correlated training data in FL for KD might compromise client privacy \cite{gong2022preserving}. Some literature has proposed alternative solutions, such as quantization \cite{sattler2021cfd} and differential privacy \cite{sattler2021fedaux}.

\subsubsection{Quantity}
KD needs a substantial amount of training data to achieve significant performance improvements \cite{liu2022communication}. This poses a challenge in some FL tasks, where collecting data can be costly. Fortunately, KD allows leveraging unlabeled datasets \cite{zhao2022semi,huang2022learn}, reducing the collection cost to some extent. However, the large amount of data requires clients to have sufficient storage, which could exclude those with limited resources. Employing a generator \cite{zhu2021data,peng2023fedgm} can effectively address the issue of collecting and storing training data for specific tasks. Furthermore, extracting knowledge from large datasets also presents challenges in terms of computation and communication bottlenecks. Dataset distillation \cite{song2023federated} offers a potential solution by attaining comparable model performance on smaller datasets.

\subsubsection{Data types}

The data employed for KD in FL can be broadly categorized into three types: local private data \cite{jeong2018communication}, public data \cite{li2019fedmd}, and synthetic data \cite{zhang2022fine}.

To use client private data for distillation, \cite{jeong2018communication} uses the average model outputs (logits) on the local dataset as a regularization term for local model loss. In \cite{yang2022fedzact}, clients upload their local models, and the server adjusts the parameters of each local model by evaluating them with each other's models, after which all local models are sent back to the clients. The clients update their local models with their private data and the predictions of other clients' models in a way similar to multi-teacher distillation. In \cite{xing2022efficient}, the server pairs all local models with each other based on the minimum mean square distance and sends the paired models to each other as their respective teacher networks. On the client side, the teacher network teaches the student network using local data and sends the student network back to the server for the next pairing round. Using client-side private datasets as the distillation dataset is based on the assumption that the model learns the personalized knowledge of a specific client, which can be reflected by its private dataset on different clients. However, this approach is not effective in non-iid scenarios because the knowledge of different clients' models is heterogeneous \cite{jeong2018communication}.

Therefore, many studies assume the availability of a public dataset \cite{lyu2023prototype,DBLP:conf/icml/ParkHH23,sattler2021cfd,gong2022preserving,yang2022fedmmd,zhao2022semi}. In \cite{sattler2021cfd}, clients and the server utilize the same public dataset for distillation. In \cite{gong2022preserving}, the logits of all client models on an unlabeled public dataset serve as knowledge to teach the global model. In \cite{yang2022fedmmd}, a labeled public dataset is utilized for mutual distillation of all client models. Specifically, clients carry out local training and send their heterogeneous models to the server. The server conducts mutual distillation on all local models using the labeled public dataset, selecting one model as the student and others as the teacher models. The distilled knowledge includes intermediate features and logits, resulting in multiple new models of a local model. Then, these models are averaged and sent back to their corresponding clients. In \cite{zhao2022semi}, the hard labels of public data samples are acquired by taking the majority vote of the hard label predictions from various client models on an unlabeled public dataset, thereby labeling the public data. Then, each client applies the global hard labels for distillation. In practice, obtaining a public dataset may be challenging, and a domain shift may exist between the public dataset and client private datasets \cite{huang2023federated,su2022domain}.

Therefore, some research \cite{zhu2021data,zhang2022dense,zhang2022fedzkt,zhang2022fine,zhang2022feddtg,qi2022fedbkd,peng2023fedgm,DBLP:conf/www/YangYPLLLZ23} has explored the feasibility of distillation without data. In \cite{zhang2022fine}, after the server aggregates all client models, it trains a generator to produce hard examples, which are samples with the highest prediction differences between local and global models. The global model then utilizes these hard examples for KD. All local models act as teachers during distillation to minimize the sum of their distillation losses. 
In \cite{zhang2022feddtg}, each client trains a generator and a discriminator locally, sends them to the server for aggregation, and obtains the global generator and discriminator. Clients use the global generator to generate fake samples, and their local private models produce soft labels on these fake samples, send them to the server for averaging, and then receive them back for distillation.

\subsection{KD-based FL: Methods}

In this section, we will summarize the main methods in KD-based FL.

Attention: In \cite{gong2021ensemble}, local model logits and attention on public datasets are uploaded to the server for distillation. Similarly, \cite{wen2023communication} uses attention-based KD to improve communication efficiency.

Continual Learning: In \cite{usmanova2022federated}, historical models and server models are used as teacher models for distillation to address the challenge of catastrophic forgetting \cite{kirkpatrick2017overcoming}. Similarly, in \cite{huang2022learn}, KD is used to learn from the previous round model and local pre-trained model to cope with catastrophic forgetting. \cite{dong2022federated} proposes a class-incremental learning mechanism to solve the catastrophic forgetting caused by new class samples.

Contrastive Learning: In \cite{han2022fedx}, contrastive learning \cite{chen2020simple} is introduced for scenarios where local data lacks labels. Contrastive loss is introduced during local distillation training. \cite{zou2023efckd} introduced model contrastive learning to improve performance.

Split Learning: In \cite{he2020group}, KD is combined with split learning \cite{vepakomma2018split}. The final model is a union of local abstract feature extractors and the server's large model. Specifically, a small model is trained locally as a feature extractor, and the client sends the output of the feature extractor as abstract features, the output of the classifier (logits), and the labels of local data to the server. The server uses abstract features, logits, and true labels for distillation to train the server-side large model and then sends the soft labels it generates back to the client. The client uses local data for distillation.

Multi-Teacher: In \cite{su2023cross}, for a special scenario where the server has a large amount of labeled data, the client sends local personalized models to the server, and the server uses these personalized models to distill into the global model. Then, the global model is sent to the client to fine-tune the local personalized model. \cite{nguyen2022label} uses a hierarchical architecture to divide all clients into multiple groups; each group selects one client as the group server and obtains a small group model through parameter sharing. Once a specific number of rounds have been completed, each small group sends the aggregated model to the global server and uses them as teacher models to perform multi-teacher distillation using the labeled public dataset on the global server.

Mutual Distillation: In \cite{wu2022communication}, the client simultaneously trains a large model and a small model, which are mutually distilled. Then, the gradients of the small model are SVD \cite{klema1980singular} decomposed and uploaded to the server. After the server reconstructs the gradients of all clients, it aggregates them and sends them to each client. \cite{li2021decentralized} is based on the P2P architecture and randomly selects two groups of clients with equal numbers, with each client corresponding to a client in the other group. The first group of clients trains the model locally and sends it to the corresponding client in the second group. The clients in the second group use local data for mutual distillation.

\bibliographystyle{ACM-Reference-Format}
\bibliography{main}